\documentclass[runningheads]{llncs}
\usepackage{graphicx}
\usepackage{comment}
\usepackage{amsmath,amssymb} 
\usepackage{color}
\usepackage{multirow}
\usepackage{subcaption}
\usepackage{adjustbox}
\usepackage{bbding}
\usepackage{array}

\def\ie{\emph{i.e. }}

\def\etal{\emph{et al. }}

\begin{document}
\pagestyle{headings}
\mainmatter
\def\ECCVSubNumber{4090}  

\title{Unsupervised Real-world Image Super Resolution via Domain-distance Aware Training} 

\titlerunning{DASR: Unsupervised Real-world Image Super Resolution} 
\authorrunning{Yunxuan Wei, Shuhang Gu, Yawei Li, Longcun Jin} 
\author{Yunxuan Wei$^{1,}$\thanks{The first two authors contribute equally to this work.}, Shuhang Gu$^{2,3,\star}$, Yawei Li$^3$, Longcun Jin$^{1,}$\thanks{Corresponding author.}}
\institute{1 South China University of Technology, 2 The University of Sydney, 3 ETH Zurich}

\maketitle
\begin{abstract}
  These days, unsupervised super-resolution (SR) has been soaring due to its practical and promising potential in real scenarios.
  The philosophy of off-the-shelf approaches lies in the augmentation of unpaired data, \ie  first generating synthetic low-resolution (LR) images $\mathcal{Y}^g$ corresponding to real-world high-resolution (HR) images $\mathcal{X}^r$ in the real-world LR domain $\mathcal{Y}^r$, and then utilizing the pseudo pairs $\{\mathcal{Y}^g, \mathcal{X}^r\}$ for training in a supervised manner.
  Unfortunately, since image translation itself is an extremely challenging task, the SR performance of these approaches are severely  limited by the domain gap between generated synthetic LR images and real LR images.
  In this paper, we propose a novel domain-distance aware super-resolution (DASR) approach for unsupervised real-world image SR.
  The domain gap between training data (e.g. $\mathcal{Y}^g$) and testing data (e.g. $\mathcal{Y}^r$) is addressed with our \textbf{domain-gap aware training} and \textbf{domain-distance weighted supervision} strategies.
  Domain-gap aware training takes additional benefit from real data in the target domain while domain-distance weighted supervision brings forward the more rational use of labeled source domain data.
  The proposed method is validated on synthetic and real datasets and the experimental results show that DASR consistently outperforms state-of-the-art unsupervised SR approaches in generating SR outputs with  more realistic and natural textures.
  Code will be available at \url{https://github.com/ShuhangGu/DASR}.
  \vspace{-2mm}
  \end{abstract}
\vspace{-5mm}
\section{Introduction}
Single image super-resolution (SR) aims at reconstructing a
high-resolution (HR) image from a low-resolution (LR) observation.
In the past two decades, SR has been a thriving research topic due to its highly practical values in enhancing image details and textures.
A wide variety of models \cite{freeman2000learning,glasner2009super,yang2010image,anrsr,cscsr} have been suggested to deal with the image SR problem. 

Benefiting from the rapid development of deep convolutional neural networks (CNNs), recent years have witnessed an explosive spread of training CNN models \cite{VDSR,EDSR,MemNet,ESRGAN,RankSRGAN,RDN,zhang2017learning,li2019_3dappearance} for SR.
State-of-the-art SR performance have been boosted by directly training networks to capture the LR-to-HR mapping.
Moreover, when combined with adversarial training \cite{Goodfellow2014GenerativeAN} or perceptual losses \cite{Johnson2016PerceptualLF}, SR networks can produce accurate and natural-looking image details.

\begin{figure}[t]
  \scriptsize
  \centering
  \begin{tabular}{p{0.5\columnwidth}<{\centering}p{0.5\columnwidth}<{\centering}}
    \begin{adjustbox}{valign=t}
      \tiny
        \begin{tabular}{l}
            \includegraphics[width=0.5\textwidth, height=0.33\textwidth]{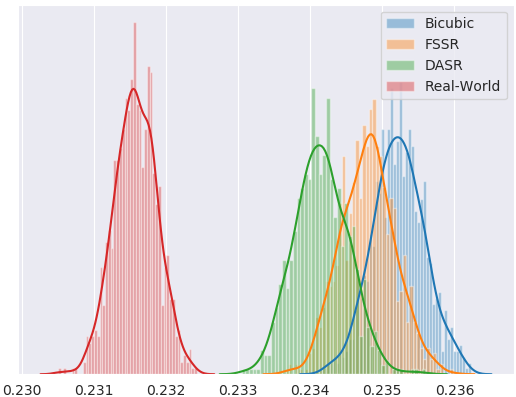}
        \end{tabular}
      \end{adjustbox}
      &
    \begin{adjustbox}{valign=t}
      \begin{tabular}{l}
          \begin{adjustbox}{valign=t}
              \tiny
              \centering
              \begin{tabular}{p{0.1\columnwidth}<{\centering}p{0.1\columnwidth}<{\centering}p{0.1\columnwidth}<{\centering}p{0.1\columnwidth}<{\centering}}
                \includegraphics[width=0.1\columnwidth,height=0.1\columnwidth]{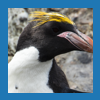} &
                \includegraphics[width=0.1\columnwidth,height=0.1\columnwidth]{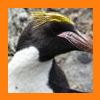} &
                \includegraphics[width=0.1\columnwidth, height=0.1\columnwidth]{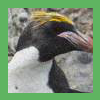} &
                \includegraphics[width=0.1\columnwidth, height=0.1\columnwidth]{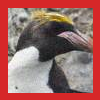} 
                \\
                \includegraphics[width=0.1\columnwidth, height=0.1\columnwidth]{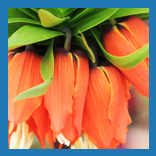} &
                \includegraphics[width=0.1\columnwidth, height=0.1\columnwidth]{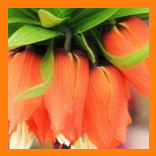}&
                \includegraphics[width=0.1\columnwidth, height=0.1\columnwidth]{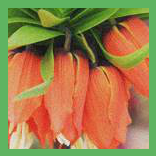}&
                \includegraphics[width=0.1\columnwidth, height=0.1\columnwidth]{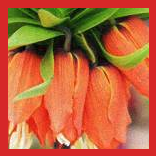}
                \\
                \includegraphics[width=0.1\columnwidth, height=0.1\columnwidth]{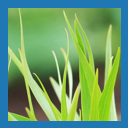} &
                \includegraphics[width=0.1\columnwidth, height=0.1\columnwidth]{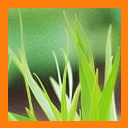} &
                \includegraphics[width=0.1\columnwidth, height=0.1\columnwidth]{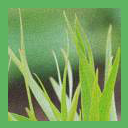}&
                \includegraphics[width=0.1\columnwidth, height=0.1\columnwidth]{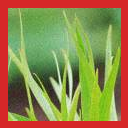}
                \\
                Bicubic&FSSR~\cite{Fritsche2019FrequencySF}&DASR(ours)&Real-world 
              \end{tabular}
          \end{adjustbox}
          \end{tabular}
  \end{adjustbox}
  \end{tabular}
  \vspace{-2mm}
  \caption{Visualization of the domain-gap between the generated LR images and the real LR images. We firstly train down-sampling networks with FSSR~\cite{Fritsche2019FrequencySF} and our DASR on the training datasets of AIM Challenge on Real World SR at ICCV 2019~\cite{lugmayr2019aim} under an unsupervised setting. Then, a discriminator is trained to separate the bicubically down-sampled data, the generated LR data by FSSR and DASR from the ground truth LR data. The histograms of the discriminator output clearly show the domain-gap between the generated and real LR images. Note that the four group of images are with the same content while the differences in these discriminator output merely come from low-level image characters.}
  \vspace{-4mm}
  \label{fig:domaingap}
\end{figure} 

In spite of their success on benchmark datasets, the poor generalization capacity of discriminatively trained SR networks limits their application in real scenarios.
When applied to super-resolve real images, SR networks trained on simulated datasets usually lead to
strange artifacts in the SR results.
For the pursuit of real-world image SR, great attempts have been made in the last several years.
By adjusting the focal length of a digital camera, several works prepared real-world SR datasets~\cite{CameraSR,Zhang2019ZoomTL,cai2019toward}.
But the collections of these datasets  is often laborious and costly.
Furthermore, SR networks trained on the collected datasets are hard to generalize to images captured in other conditions.
Another category of approaches investigate real-world image SR from an algorithmic perspective.
Some works~\cite{michaeli2013nonparametric,Kernel_modeling,bell2019blind} assume the LR and HR images satisfy a parameterized degradation model, and propose  blind SR algorithms which are able to adapt to the unknown down-sampling kernel in the testing phase.
These blind SR algorithms \cite{cornillere2019blind,Yuan2018UnsupervisedIS,SRMD,ZSSR} have shown improved generalization capacity over models trained on predetermined synthetic data, but the fixed degradation assumption greatly limits their performances on real data, which are often subject to complex sensor noise and compression artifacts.

Recently, without any assumption on the degradation model, unsupervised SR algorithms have been proposed to leverage unpaired training data.
Given a set of real-world LR images $\mathcal{Y}^r=\{y^r_i\}_{i=1,\dots,N}$, some works proposed to train a degradation network to generate LR observations $y^g_i$ of the  available HR images $x^r_i\in \mathcal{X}^r$, and enforcing the same distribution of the generated LR images $\mathcal{Y}^g=\{y^g_i\}_{i=1,\dots,M}$ with that of real LR images $\mathcal{Y}^r$~\cite{To_learn,lugmayr2019unsupervised,Fritsche2019FrequencySF}.
With the generated pseudo pairs $\{\mathcal{Y}^g, \mathcal{X}^r\}$, supervised training can be employed to train the SR network.
Such unsupervised settings exploit the real training data to learn the complex degradation model and lead to promising SR results on real-world images \cite{lugmayr2019aim}.
However, existing unsupervised SR approaches \cite{Fritsche2019FrequencySF,To_learn,lugmayr2019unsupervised} ignored the domain-gap between $\mathcal{Y}^g$ and $\mathcal{Y}^r$ in the training process of SR networks.
In Fig. \ref{fig:domaingap}, we present a visual example to show the domain gap between the generated and the real LR images.
Although the trained down-sampling networks are able to generate better LR images that resides in a domain closer to the real LR domain than the bicubically downsampled images, domain gap still exists between $\mathcal{Y}^g$ and $\mathcal{Y}^r$.

In this paper, we propose a Domain-distance Aware Super-resolution (DASR) framework for real-world image super-resolution.
Different from previous unsupervised methods \cite{To_learn,lugmayr2019unsupervised,Fritsche2019FrequencySF,han2019unsupervised} which rely on the generation of pseudo pairs for supervised training, our DASR takes into consideration the domain gap between the generated and real LR images, \ie $\mathcal{Y}^g$ and $\mathcal{Y}^r$, and solves the SR problem with both of them under a domain adaptation setting.
Our DASR method addresses the domain gap issue through two training strategies:  {\bf \textit{domain-gap aware training}} and {\bf \textit{domain-distance weighted supervision}}.

\textbf{Firstly}, with the domain-gap aware training, DASR employs both the generated pseudo pairs $\{\mathcal{Y}^g, \mathcal{X}^r\}$ and real LR images $\mathcal{Y}^r$ to train the SR network.
Besides the supervised loss on the pseudo pairs $\{\mathcal{Y}^g, \mathcal{X}^r\}$, DASR also imposes adversarial constraints on the HR estimation $\hat{\mathcal{X}}^{r\rightarrow r}$ 
of real-world data $\mathcal{Y}^r$.
Incorporating $\mathcal{Y}^r$ into training informs the network of the target domain, greatly improves its SR performance on real-world data.

\textbf{Secondly}, besides the domain-gap aware training, a domain-distance weighted supervision strategy is also proposed for an advanced exploitation of the generated pseudo pairs.
As shown in Fig. \ref{fig:domaingap}, some generated LR samples reside closer to the real-world domain while the others are relatively far away from it.
We therefore adjust the importance of each  pair $\{y^g_i, x^r_i\}$ according to the domain distance between $y^g_i$ and $\mathcal{Y}^r$.
Samples which are relatively closer to the real-world domain are assigned with larger weights in the training phase;
while unrealistic samples are only allowed to make a limited contribution for the training.

In addition to the above strategies, which are the major contributions of this paper, we also improve previous methods by employing better architecture of down-sampling network and better adversarial loss in the wavelet domain.

Our contributions can be summarized as follows:
\begin{itemize}
  \item A domain distance aware super-resolution (DASR) framework is proposed to solve the real-world image SR problem. DASR addresses the domain gap between generated LR images and real images with the proposed domain-gap aware training and domain-distance weighted supervision strategies.
  \item We provide detailed ablation studies to analyze and
validate our contributions. Experimental results on
synthetic and real datasets clearly demonstrate the superiority of DASR over the competing approaches.

  \end{itemize}

\section{Related Works}
\subsection{Single Image Super-Resolution with CNNs}
CNN-based SR methods have dominated the single image SR field in the last several years.
In the pioneer work \cite{Dong2014LearningAD}, Dong \etal proposed the first CNN-based SR method, \ie SRCNN, which utilizes a 3 layers CNN to directly learn the mapping function between LR and HR image pairs.
Since that, an increasing number of researchers  ventured into the
deep learning field.
A surge of network architectures, such as a deep network with residual learning \cite{VDSR}, network with residual blocks \cite{EDSR}, densely connected network \cite{ESRGAN}, have been designed to solve the SR task.
The SR performance on benchmark datasets have been continuously improved by newly proposed network architectures \cite{Ahn2018FastAA,Fan2017BalancedTR,Haris2018DeepBN,Huang2018DenselyCH,Kim2015AccurateIS,Lai2017DeepLP,Dong2014ImageSU,MemNet,RDN,ahn2018image,huang2015single,kim2016accurate,tong2017image}.  
Besides investigating more powerful network architecture, perceptual-driven approaches 
investigate better loss functions to improve the perceptual quality of SR results.
Johnson \etal \cite{Johnson2016PerceptualLF} proposed a perceptual loss which  measures the error of two images in the feature space instead of pixel space.
Ledig \etal \cite{SRGAN} firstly introduced the adversarial loss 
 to favor outputs residing on the manifold of natural images.
Inspired by these pioneer works, different training criterions \cite{ESRGAN,RankSRGAN,enhancenet} have been suggested to promote visual quality of SR results.

Although significant advances have been made, all the aforementioned approaches are trained and evaluated on simulated datasets which assume simple and uniform degradation.
These days, the real-world image super-resolution problem has attracted increasing attentions due to its high practical values.
A branch of work~\cite{michaeli2013nonparametric,Kernel_modeling,bell2019blind,cornillere2019blind} assumes the degradation model between LR and HR images can be characterized by an unknown blur kernel and the subsequent downsampling operation.
These blind SR works explicitly estimate the unknown blur kernel at the testing time, and take the estimated kernel as input variable for kernel adaptive SR networks \cite{SRMD,Kernel_modeling} to adapt to different degradation hyper-parameters.
There are also works ~\cite{ZSSR,Liang2017SingleIS} attempt to use the information of testing image for training or fine-tuning the SR network in the testing phase.
However, both approaches
still rely on a known degradation model during training.
To deal with more general real-world SR task, some recent works consider an unsupervised setting which does not rely on the degradation assumption.
Given a group of  LR images, Yuan et al.~\cite{Yuan2018UnsupervisedIS}  firstly learned a mapping to transfer the original input images to the clean image domain, and applied SR in the clean image domain.
Other unsupervised approaches ~\cite{Fritsche2019FrequencySF,To_learn,lugmayr2019unsupervised} proposed to learn a downsampling process to generate paired data, and train SR network with the generated data in a supervised manner.
The advantage of these unsupervised SR methods is that they do not rely on the degradation assumption, and therefore are capable of generalizing to very challenging real-world images.
However, as image translation itself is an extremely challenging task, the generated LR images are often not consist with the character of real LR images.
Such domain gap between training and testing data will deteriorate the final SR performance in the testing phase.
\subsection{Domain Adaptation}
    \vspace{-2mm}
Domain adaptation aims to utilize a labeled source
domain to learn a model that performs well on an unlabeled
target domain.
It is a classical machine learning problem~\cite{ganin2014unsupervised,fernando2013unsupervised,sener2016learning}.
Recently, with the explosive spread of using CNN models to solve computer vision tasks, domain adaption has received an increasing attention. It has been deployed in many tasks for levering synthetic data or data from other datasets.
Early domain adaptation works in computer vision field  focus on solving the domain bias issue in high-level classification  tasks \cite{ren2018adversarial,Chen2018DomainAF,ghifary2016deep,gopalan2011domain,saito2018maximum}.
While recently,  domain adaptation has also been adopted in more challenging dense estimation tasks such as semantic segmentation ~\cite{zhang2017curriculum,sankaranarayanan2017unsupervised}.
With appropriate adaptation strategies, models trained on synthetic datasets have achieved comparable performance 
to models trained with real labeled data \cite{bousmalis2017unsupervised,ganin2014unsupervised,sankaranarayanan2017unsupervised,MDLCC}.
In this paper, we utilize domain adaptation to improve SR performance on real data.
To the best of our knowledge, our work is the first attempt of exploiting  domain adaptation  to improve  low-level image enhancement performance.

    \vspace{-1mm}
\section{DASR for Unsupervised Real-World Image SR}
    \vspace{-1mm}
\subsection{Methodology Overview}\label{sec:Methods}
    \vspace{-1mm}
\begin{figure}[t!]
  \centering
  \includegraphics[scale=0.175]{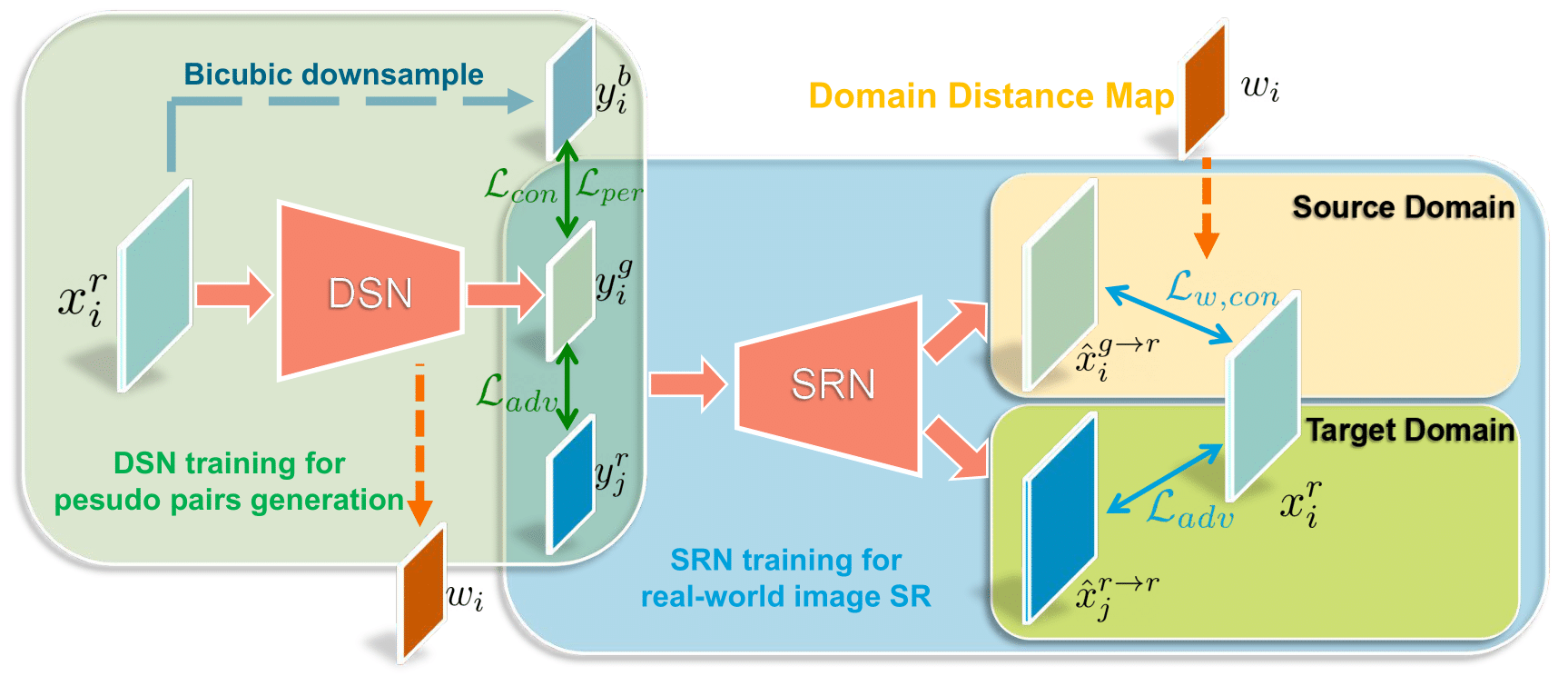}
      \vspace{-3mm}
  \caption{Illustration of our DASR framework.}
  \label{fig:illustration}
    \vspace{-5mm}
\end{figure}
In this paper, we study the unsupervised real-world image SR problem.
Given a set of real-world LR images $\mathcal{Y}^r=\{y^r_i\}_{i=1,\dots,N}$ and a set of unpaired HR images $\mathcal{X}^r=\{x^r_i\}_{i=1,\dots,M}$, we aim to learn a SR network (SRN) to enlarge the size of LR image and simultaneously ensure the HR estimation lies in the real HR distribution.
To attain this goal, we follow the previous state-of-the-art methods ~\cite{Fritsche2019FrequencySF,To_learn,lugmayr2019unsupervised} and propose a two-stage approach.
Firstly, we train a  down-sampling network (DSN) to generate LR images in the real-world LR domain from HR images: $y^g_i=DSN(x^r_i)$.
Then, we utilize the generated LR-HR pairs $\{y^g_i, x^r_i\}_{i=1,\dots,M}$ for training the SRN.
In contrast to previous works \cite{Fritsche2019FrequencySF,To_learn,lugmayr2019unsupervised} which simply employ the generated pseudo pairs to train SRN in a supervised manner,
our DASR framework considers the domain bias between $\mathcal{Y}^g$ and $\mathcal{Y}^r$ and adopts domain-gap aware training and  domain-distance weighted supervision strategies to take full advantage of real-world LR images as well as the generated pairs. An illustration of the proposed DASR framework is shown in Fig. \ref{fig:illustration}.

In the remaining parts of this section, we firstly introduce how we train our DSN network to generate synthetic LR-HR pairs.
Then, we present our domain-gap aware training strategy and domain-distance weighted supervision strategy.

\vspace{-2mm}
\subsection{Training of Down-Sampling Network}
In this section, we introduce how we train a Down-Sampling Network (DSN) from unpaired data to generate synthetic LR-HR pairs.

\vspace{2mm}

\noindent\textbf{Network architecture.}
Different down-sampling networks \cite{Fritsche2019FrequencySF} have been trained in previous unsupervised SR works to generate  synthetic real-world LR  images from HR images.
To avoid changing the image sizes between input to output, existing approaches adopt a bicubic downsampling operation as the pre-processing step.
Therefore, the degradation networks only need to translate bicubic downsampled image to the real image domain.
Despite reducing the difficulty in translation, the pre-processing downsampling operation may remove necessary information for  generating real LR images.
In contrast, our DSN takes HR image as input and captures  the  whole  degradation  process with the network directly.
 Without  losing  information  in  the bicubic down-sampling step, all the information in HR images can be exploited for generating better synthetic LR images.
Our detailed network architecture can be found in Fig. 3 (a).
DSN utilizes 23 residual blocks to extract information from the HR image, each residual block contains two convolutional
layers (with kernel size 3$\times$3 and channel number 64) and a ReLU activations in between.
Then, a bilinear resize operator and two convolutional layers are adopted to reduce the spatial resolution of features and project the features back to the image domain.

\begin{figure}[t!]
  \centering
  \begin{tabular}{p{0.5\columnwidth}<{}p{0.5\columnwidth}<{}}
    \includegraphics[scale=0.18]{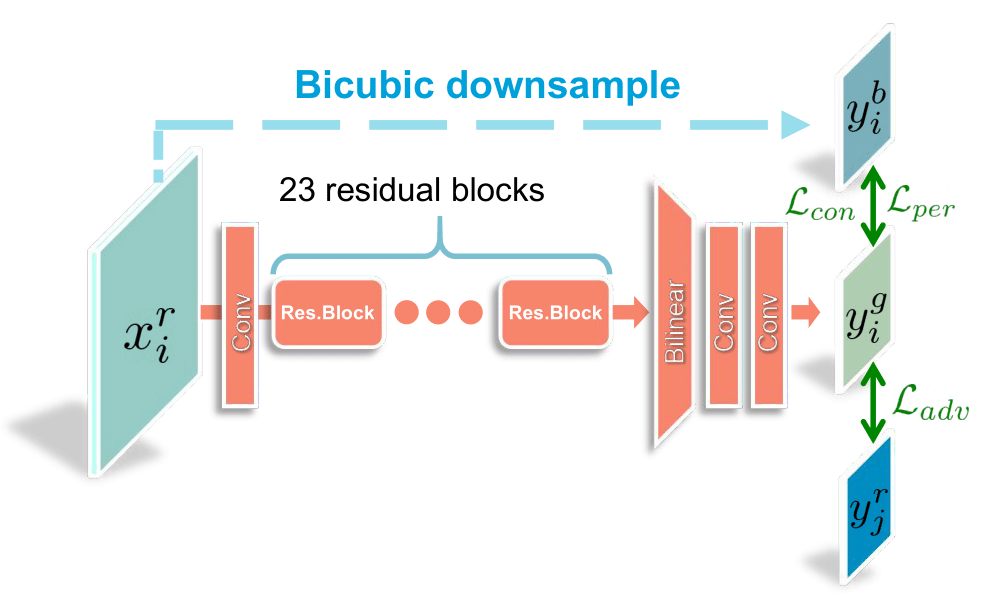}&\includegraphics[scale=0.14]{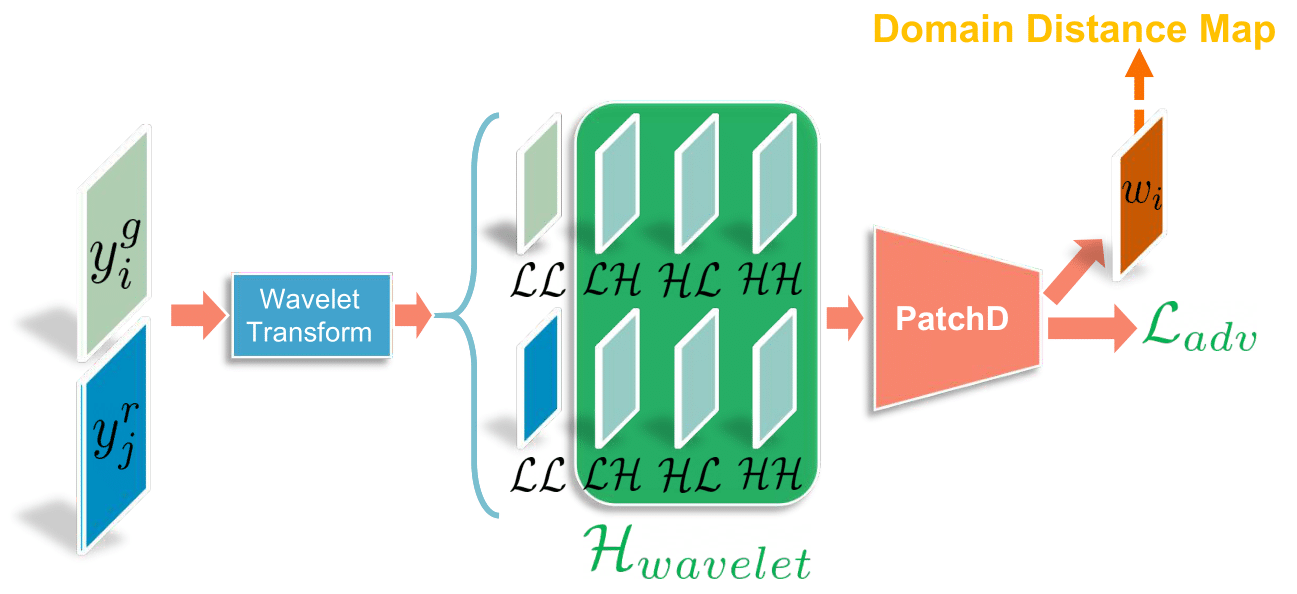} \\
    \centering(a) & \centering(b) \\
  \end{tabular}
    \vspace{-7mm}
  \caption{(a) Network architecture of our down-sampling network. (b) Adversarial loss
  in the wavelet high-frequency space.}
  \label{fig:DSN}
  \vspace{-1mm}
\end{figure}

\vspace{2mm}
\noindent\textbf{Losses.}
We train our DSN with a combination of multiple loss functions.
To keep the content of generated LR image consistent with the input HR image, we apply content loss $\mathcal{L}_{con}$ and perceptual loss $\mathcal{L}_{per}$ to constrain the distance between generated LR image $y_i^g=DSN(x_i^r)$ and bicubic downsampled HR image $y_i^b$:
\begin{equation}
  \mathcal{L}_{con} = \mathbb{E}_{x^r}||y_i^b-DSN(x_i^r)||_1,~~~
  \mathcal{L}_{per} =
  \mathbb{E}_{x^r}\|\phi(y_i^b) - \phi(DSN(x_i^r))\|_1;
\end{equation}
where $y_i^b=\mathcal{B}(x_i^r)$ is the bicubic downsampled HR image, and $\phi(\cdot)$ denotes the VGG \cite{VGG} feature extractor.
In our implementation, we follow ESRGAN \cite{ESRGAN} and calculate perceptual loss on VGG-19 \cite{VGG} features from \textit{conv5\_3} convolutional layer.
While, to achieve the goal of domain translation, we impose adversarial losses between image samples in $\mathcal{Y}^g$ and $\mathcal{Y}^r$.
We adopt a similar idea with FSSR \cite{Fritsche2019FrequencySF}, which only imposes adversarial loss in the high-frequency space.
But we use Haar wavelet transform to extract more informative high-frequency component.
Concretely, denote the four sub-bands decomposed by Haar wavelet transform as $\mathcal{LL}$, $\mathcal{LH}$, $\mathcal{HL}$ and $\mathcal{HH}$, we stack the $\mathcal{LH}$, $\mathcal{HL}$ and $\mathcal{HH}$ components as the input to the discriminator.
Compared with the high-frequency extractor used in FSSR~\cite{Fritsche2019FrequencySF}, our wavelet-based extractor also exploits direction information to better characterize image details.
The GAN loss for generator (e.g. our DSN) is defined as:
\begin{equation}
  \mathcal{L}_{adv}^G = \mathbb{E}_{x^r}[log\left(1-D\left(\mathcal{H}_{wavelet}\left(DSN(x^r)\right)\right)\right)];
\label{eq:g1}
\end{equation}
and the GAN loss for training the discriminator is in a symmetrical form:
\begin{equation}
  \mathcal{L}_{adv}^D = \mathbb{E}_{y^r}[log(1-D(\mathcal{H}_{wavelet}(y^r)))]+\mathbb{E}_{x^r}[log\left(D\left(\mathcal{H}_{wavelet}\left(DSN(x^r)\right)\right)\right)].
\label{eq:d1}
\end{equation}
$\mathcal{H}_{wavelet}(\cdot)$ in Eq. \ref{eq:g1} and \ref{eq:d1} represents extracting $\mathcal{LH}$, $\mathcal{HL}$ and $\mathcal{HH}$ subbands with Haar wavelet transform and concatenating the three variables.
Imposing the adversarial loss in the high-frequency domain enables us to ignore the low-frequency content which is less relevant to the SR task \cite{Fritsche2019FrequencySF} and focus more on the image details.
Moreover, conducting adversarial training in lower-dimension space also reduces the difficulty of GAN training \cite{Knop2019SlicedGM,wu2019sliced}.

In our implementation, we adopt a similar strategy as CycleGAN \cite{CycleGAN2017}, which impose GAN loss on each patches.
Concretely, we utilize a 4 layer fully convolutional discriminator, the patch discriminator has a valid receptive field of $23 \times 23$.
The PatchGAN strategy helps to derive the patch-level dense domain distance map, which will be utilized in the subsequent training phase of SRN.
More details of our patch discriminator can be found in our supplementary file.

\vspace{3mm}
\noindent\textbf{Training Details.}
Our DSN is trained with  the following losses:
\begin{equation}
  \mathcal{L}_{DSN} = \alpha \mathcal{L}_{con} + \beta \mathcal{L}_{per} + \gamma \mathcal{L}_{adv}^G.
  \label{eq:lossdsn}
\end{equation}
To stabilize our training, we pre-train our DSN network with the content loss.
After a pre-train process of 25000 iterations, the $\alpha$, $\beta$ and $\gamma$ in Eq. \ref{eq:lossdsn} are set as 
0.01, 1 and 0.0005, respectively.
We train the DSN networks with 192 $\times$ 192 HR crops, the batch size is set as 16.
The initial learning rate is 0.0001, and we halve the learning rate every 10000 iterations.
We train the model for 50000 iterations.

\vspace{-2mm}
\subsection{Domain distance aware training of Super-Resolution Network}
\begin{figure}[t!]
  \centering
  \includegraphics[scale=0.175]{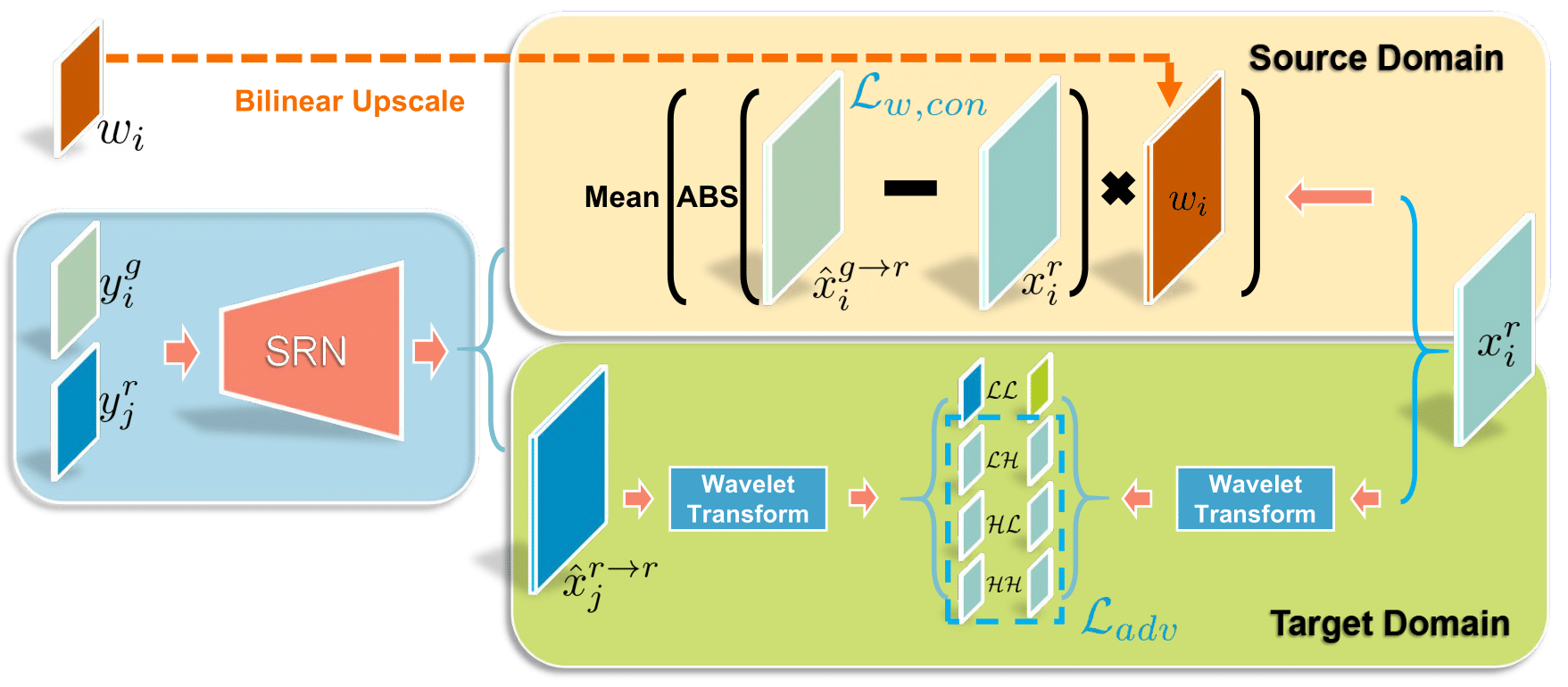}
      \vspace{-3mm}
  \caption{Domain-distance aware training of our SRN.}
  \label{fig:SRN}
    \vspace{-3mm}
\end{figure}

With the aforementioned DSN, we are able to generate synthetic paired data $\{y_i^g, x_i^r\}_{i=1,\dots,M}$. 
 However, as we have validated in Fig. \ref{fig:domaingap}, domain gap still exists between generated LR images $\mathcal{Y}^g$ and real LR images $\mathcal{Y}^r$.
 When the SR network trained on synthetic data is applied to super-resolve real-world LR images,
 such domain gap between training and testing data will lead to a performance drop.
To alleviate the domain bias issue, we consider a domain adaptation setting and incorporate both source domain labeled data $\{\mathcal{Y}^g, \mathcal{X}^r\}$ and target domain unlabeled data $\mathcal{Y}^r$ in training of our SR network.
The core of our adaptation strategy consists of two parts which include domain-gap aware training and domain-distance weighted supervision. An illustration of the domain-distance aware training process is presented in Fig. \ref{fig:SRN}.

\vspace{2mm}
\noindent\textbf{Domain-gap aware training.}
Given training samples from the source and the target domains, we utilize different losses in the two domains to take full advantage of the training data.
For the data in the source domain, which have supervised labels, we deploy losses to train the network in a supervised manner.
While, for the data in the target domain, which do not have labels, we impose adversarial losses to align the distribution of their outputs $\hat{\mathcal{X}}^{r\rightarrow r}=SRN(\mathcal{Y}^{r})$ and the distribution of real HR images $\mathcal{X}^{r}$.
The same as our DSN training, we introduce GAN losses in the wavelet space.
\begin{equation*}
\vspace{-1mm}
\begin{aligned}
  \mathcal{L}_{target,adv}^G = &   \mathbb{E}_{y^r}[log(\!1-\!D(\mathcal{H}_{wavelet}(SRN(y^r))))];\\
  \mathcal{L}_{target,adv}^D = & \mathbb{E}_{x^r}[log(1\!-\!D(\mathcal{H}_{wavelet}(x^r)))]\!+\!\mathbb{E}_{y^r}[log(D(\mathcal{H}_{wavelet}(SRN(y^r))))].
  \end{aligned}
  \vspace{-1mm}
\end{equation*}
Besides introducing $\mathcal{L}_{target,adv}$ to guide the network training with target domain information, making rational use of information in the source domain is of equally importance for obtaining a good SRN.
In the following part, we introduce how the domain distance information of each sample can be utilized for adaptively supervise the training of SRN.

\vspace{2mm}
\noindent\textbf{Domain-distance weighted supervision.}
As shown in Fig. \ref{fig:domaingap}, each sample in $\mathcal{Y}^g$ has distinct distance to the real-world image domain $\mathcal{Y}^r$.
More Specifically, since the difference between images from different domains only lies in their low-level characters, each area of generated images may possess diverse domain distance to the real-world image domain.
When being applied as source domain data to train target domain SRN, different areas should be endowed with various 
importance based on their respective distance to target domain.
We therefore propose a weighted supervision strategy which utilizes dense domain distance map to adaptively adjust the losses for each  pair $\{y^g_i, x^r_i\}$.
The weighted supervised losses in the source domain can be written as follows:
\begin{equation}
\vspace{-2mm}
  \begin{aligned}
\mathcal{L}_{source, con} &= E_{y_i^g,x_i^r}\|w_i\odot\left((SRN(y_i^g)-x_i^r\right)\|_1,\\
\mathcal{L}_{source, per} &= E_{y_i^g,x_i^r}\|w_i\odot(\phi\left(SRN(y_i^g))-\phi(x_i^r)\right)\|_1;
\end{aligned}
\vspace{-1mm}
\end{equation}
where $w_i$ is the domain distance map for $y_i^g$, and $\odot$ denotes the point-wise multiplication. 
We utilize the discriminators obtained during the training process of DSN to evaluate the domain distance map for each sample.
Note that the discriminator is trained to distinguish the generated patches from the real-world LR patches and the discriminator output denotes the possibility that the input comes from the target domain.
Thus, the larger the discriminator output, the higher the possibility that the input comes from the target real-world LR domain and the less the distance to the target domain.
We directly utilize the discriminator outputs to weight the importance of each local area.
We utilize bilinear resize to adjust the weight map size, make it consistent with the HR image.

\vspace{2mm}
\noindent\textbf{Training details.}
In  summary, with our domain-distance aware training strategy, SRN is trained through minimizing the following losses:
\begin{equation}
\vspace{-1mm}
  \begin{aligned}
\mathcal{L}_{SRN} &= \alpha\mathcal{L}_{source, con} + \beta\mathcal{L}_{source, per}  + \gamma\mathcal{L}_{target, adv}.\label{eq:srnloss}
\end{aligned}
\vspace{-1mm}
\end{equation}
The same as our training schedule for DSN, we pretrain our SRN with content loss in source domain.
After 25000 iterations of pretraining, we employ all the losses in Eq. \ref{eq:srnloss} with weights $\alpha=0.01$, $\beta=1$ and $\gamma = 0.005$ to train the network for another 50000 iterations.
We initialize the learning rate as 0.0002, and halve it every 10000 iterations.

Our adaptation strategy is applicable to diverse network architectures.
In this paper, we directly adopt the architecture used in ESRGAN \cite{ESRGAN} as our SRN. 

\vspace{-3mm}
\section{Experimental Results on Synthetic Datasets}
\subsection{Experimental Setting}\label{syn_ex_setting}
In this section, we evaluate the proposed DASR method on the AIM dataset, which was used in the AIM Challenge on Real World SR at ICCV 2019~\cite{lugmayr2019aim}.
The dataset was simulated by applying synthetic but realistic degradations to clean high-quality images.
We follow the experimental setting of \textit{target domain super resolution} in the Challenge.
The training set consists of 2650 noisy and compressed images with unknown degradation from the Flickr2K dataset \cite{Flickr2K}, and 800 clean HR images from the DIV2K \cite{Timofte_2018_CVPR_Workshops} dataset.
We conduct our experiments on the validation dataset of the AIM challenge, which has paired data for quantitative comparison.
The validation dataset contains 100 images with the same type of degradation as the training LR images.
Since the GAN approaches focus on the perceptual quality of the recovered image, Learned Perceptual Image Patch Similarity (LPIPS) and Mean Opinion Score (MOS) are used as the primary metrics to evaluate different methods.
A user study is conducted to calculate the MOS for different methods.
The test candidates were shown a side-by-side comparison of a sample result and the corresponding ground-truth. 
The final MOS of a specific image is the average score of  different candidates' opinion: 
 0 - `the same', 1 - `very similar', 2 -
`similar', 3 - `not similar' and 4 - `different'.
For all the MOS values reported in the paper, we have the same 26 candidates
to perform the user study.
In addition to the perceptual metrics, the Peak Signal-to-Noise Ratio (PSNR) and Structural Similarity index (SSIM) are used as additional quantitative metrics for the reference.

\vspace{-2mm}
\subsection{Ablation study}\label{sec:ablations}
Before comparing DASR with state-of-the-art unsupervised real-world image SR methods, we conduct ablation experiments to analyze our DASR model.
We firstly analyze our design choice for DSN training.
Then, we provide experimental results to demonstrate the effectiveness of the proposed Domain-gap aware training and Domain-distance weighted supervision strategies.

\begin{table}[b!]
\vspace{-3mm}
\caption{LR image generation with different DSN architectures and adversarial training criterion. Details of the experiments are described in \ref{sec:ablations}.}
\label{table:ablation1}
    \centering   
    \begin{tabular}{cc}
      \centering   
      \begin{tabular}{c|c|ccc}
      \hline\hline
        DSN Input & $\mathcal{L}_{adv}$ for DSN & LPIPS & PSNR & SSIM \\
        \hline
        Bicubic LR & GBFS & 0.110 & 25.258 & 0.8081 \\
        HR & RGB & 0.138 & 25.204 & 0.8153 \\
        HR & GBFS & 0.101 & 25.474 & 0.8156 \\
        HR & WFS & 0.067 & 26.007 & 0.8097 \\
       \hline\hline
      \end{tabular}
      &
      \tiny\hspace{-1mm}
      \begin{tabular}{p{0.08\columnwidth}<{\centering}p{0.08\columnwidth}<{\centering}p{0.08\columnwidth}<{\centering}p{0.08\columnwidth}<{\centering}p{0.08\columnwidth}<{\centering}}
        \centering
        \includegraphics[scale=0.32]{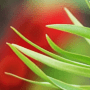}&
        \includegraphics[scale=0.32]{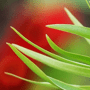} &
        \includegraphics[scale=0.32]{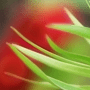} &
        \includegraphics[scale=0.32]{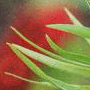} &
        \includegraphics[scale=0.32]{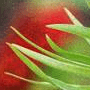} 
        \\
        \includegraphics[scale=0.32]{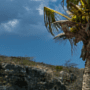} &
        \includegraphics[scale=0.32]{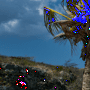} &
        \includegraphics[scale=0.32]{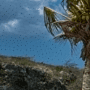} &
        \includegraphics[scale=0.32]{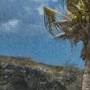} &
        \includegraphics[scale=0.32]{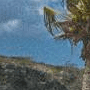}
        \\
         Bicubic& HR &HR& HR&Real\\
         GBFS&RGB & GBFS&WFS&world 
      \end{tabular}
    \end{tabular}
    \vspace{-5mm}
\end{table}

\begin{table}
    \vspace{-2mm}
    \caption{Ablation study on the AIM dataset \cite{lugmayr2019aim}. We evaluate our \textit{Domain gap aware training} and \textit{Domain distance weighted supervision} strategies in different conditions. Details of the experiments are described in \ref{sec:ablations}.}
    \scriptsize
    \begin{center}
        \label{table:ablation2}
        \begin{tabular}{p{0.20\columnwidth}<{\centering}|p{0.123\columnwidth}<{\centering}|p{0.123\columnwidth}<{\centering}|p{0.123\columnwidth}<{\centering}|p{0.123\columnwidth}<{\centering}|p{0.123\columnwidth}<{\centering}|p{0.123\columnwidth}<{\centering}}
        \hline
        \hline
        Source domain&~~$\{\mathcal{Y}^b,\mathcal{X}^r\}$~~&~~ $\{\mathcal{Y}^b,\mathcal{X}^r\}$~~& ~~$\{\mathcal{Y}^g,\mathcal{X}^r\}$ ~~&~~$\{\mathcal{Y}^g,\mathcal{X}^r\}$~~& ~~~$\{\mathcal{Y}^g,\mathcal{X}^r\}$~~&~~$\{\mathcal{Y}^g,\mathcal{X}^r\}$~~\\
        Target domain& ~~~- ~~~ &  ~~~$\mathcal{Y}^r$ ~~~&  ~~~- ~~~& ~~~$\mathcal{Y}^r$ ~~~&  ~~~- ~~~& ~~~$\mathcal{Y}^r$~~~\\
        Domain gap aware& \XSolidBrush& \CheckmarkBold& \XSolidBrush&\CheckmarkBold& \XSolidBrush& \CheckmarkBold\\
        Weighted sup.& \XSolidBrush& \XSolidBrush& \XSolidBrush&\XSolidBrush&\CheckmarkBold& \CheckmarkBold\\
        \hline
        PSNR & 21.382 & 20.820 &21.910&21.805&21.452&21.600\\
        SSIM & 0.5478 & 0.5103 &0.5555&0.5615&0.5304&0.5640\\
        LPIPS & 0.543 & 0.390 &0.378&0.359&0.348&0.336\\
        MOS & 3.16&2.87&2.41&2.37&2.31&1.94 \\
        \hline
        \hline
        \end{tabular}
    \end{center}

    \vspace{-3mm}
    \begin{tabular}{cc}
        \begin{adjustbox}{valign=c}
        \includegraphics[height=0.14\columnwidth, width=0.193\columnwidth]{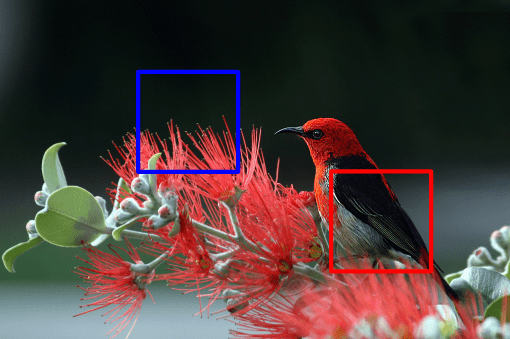}
        \end{adjustbox}
        &
        \begin{adjustbox}{valign=c}
            \begin{tabular}{p{0.125\columnwidth}<{\centering}p{0.125\columnwidth}<{\centering}p{0.125\columnwidth}<{\centering}p{0.125\columnwidth}<{\centering}p{0.125\columnwidth}<{\centering}p{0.125\columnwidth}<{\centering}}
            
            \includegraphics[width=0.13\columnwidth]{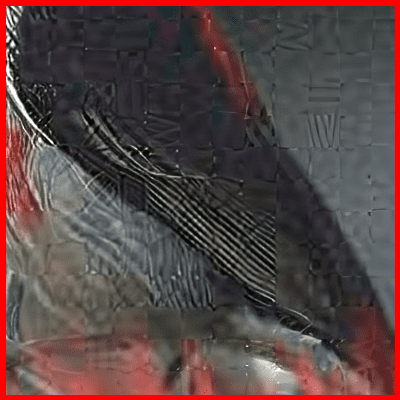} &
            \includegraphics[width=0.13\columnwidth]{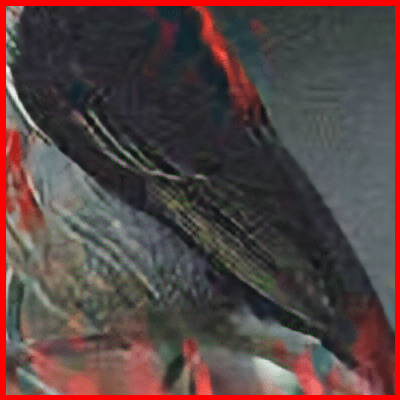} &
            \includegraphics[width=0.13\columnwidth]{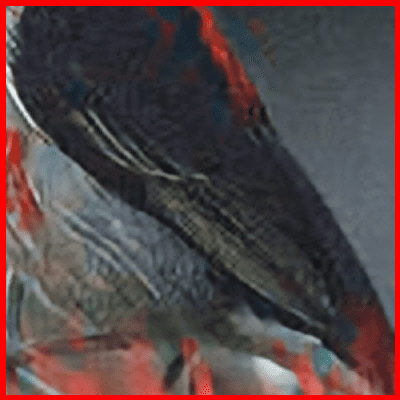}&
            \includegraphics[width=0.13\columnwidth]{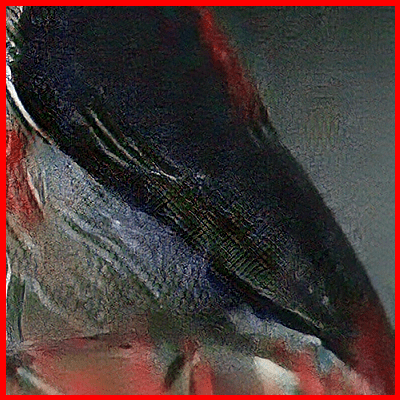} &
            \includegraphics[width=0.13\columnwidth]{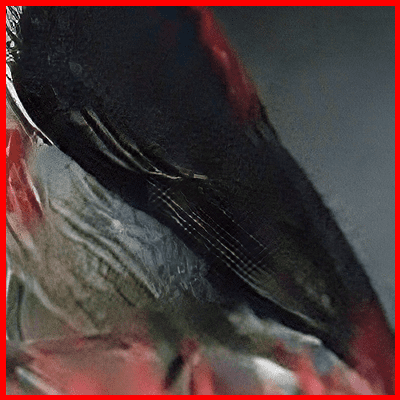} &
            \includegraphics[width=0.13\columnwidth]{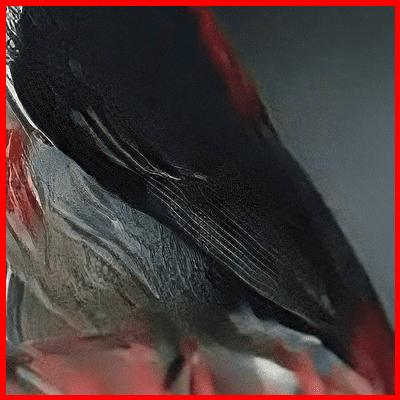} \\
            \includegraphics[width=0.13\columnwidth]{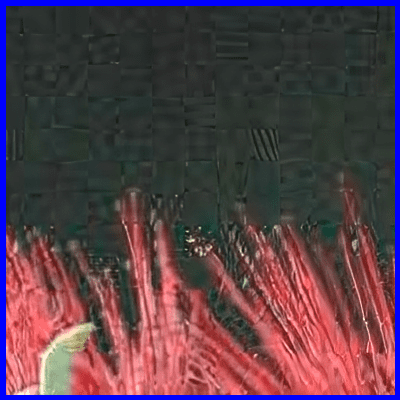} &
            \includegraphics[width=0.13\columnwidth]{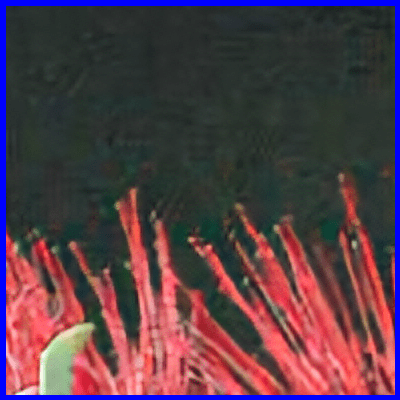} &
            \includegraphics[width=0.13\columnwidth]{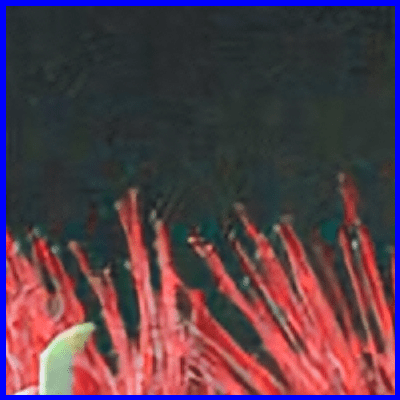} &
            \includegraphics[width=0.13\columnwidth]{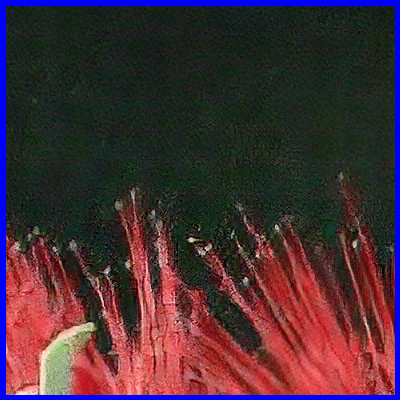} &
            \includegraphics[width=0.13\columnwidth]{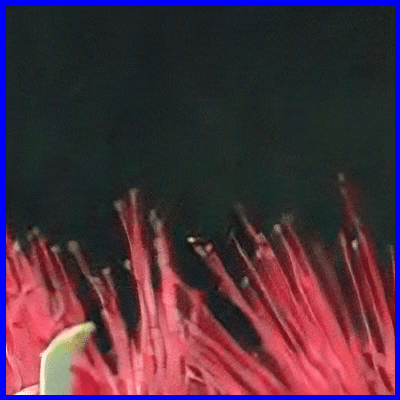}&
            \includegraphics[width=0.13\columnwidth]{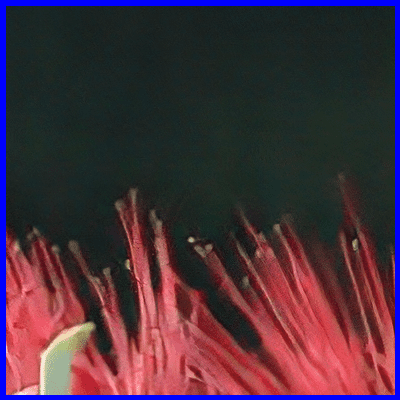}
            \end{tabular}
        \end{adjustbox}
    \end{tabular}
    \vspace{-6mm}
\end{table}

\vspace{2mm}
\noindent\textbf{Better down-sampling network for synthetic paired data generation.}
The idea of training a down-sampling network has been explored in previous approaches~\cite{Fritsche2019FrequencySF,To_learn,lugmayr2019unsupervised}.
Our DSN improves previous methods by directly estimating LR image from un-preprocessed HR image and adopting better adversarial loss in wavelet space.
In order to evaluate the effect of our modifications, we combine different choices of design to generate LR images.
We train downsampling networks with different settings, and use these models to generate LR images from the HR images in AIM validation dataset.
We compare the generated LR images with the original LR images in the datasets,
the quantitative metrics achieved by different models are reported in Table 1.
Some visual examples of the LR images generated by different downsampling networks are also presented.

In the table, HR/Bicubic LR denote the respective inputs used by different down-sampling networks.
While, Gaussian Blur Frequency Separation (GBFS), Wavelet Frequency Separation (WFS) and RGB indicate the model conducts adversarial training in different spaces: GBFS uses the residual between original and Gaussian blurred images to extract high frequency component, our WFS approach adopts Wavelet transform to obtain high frequency component, RGB means we introduce GAN loss directly on RGB images.
The results in Table 1 clearly show that both the proposed architecture of DSN and adversarial loss in the wavelet space are beneficial for generating better LR images, which are more similar to the real images in the target domain.

\vspace{-2mm}
\noindent\textbf{Domain-gap Aware Training.}
As one of our major contributions, domain-gap aware training is of vital importance for the success of our model.
In Table 2, we present experimental results to show the advantage of our domain-gap aware training.
We conduct experiments with our DSN generated synthetic pairs $\{\mathcal{Y}^g, \mathcal{X}^r\}$ or bicubic downsampled LR-HR pairs $\{\mathcal{Y}^b, \mathcal{X}^r\}$ as source domain data.
As we have introduced in section 3.2, our domain-gap aware training strategy introduces extra adversarial loss in the target domain.
For fair comparison, in table 2, the models  without domain-gap aware training introduce the adversarial loss in the source domain.
The same strategy has been widely adopted in previous unsupervised methods ~\cite{ESRGAN,SRGAN,RankSRGAN}.
In both settings, the proposed domain-aware training  strategy consistently improves the final SR performance.
It helps the SRN to generate high quality HR estimations with better MOS score as well as better LPIPS index.
Because the MOS index is achieved by comparing subject image with its corresponding reference image, images with different visual quality may be categorized as the same class, i.e. similar or not similar. 
Therefore, the MOS score could not thoroughly reflect the advantage of our domain-gap aware training strategy.
On the other hand, the LPIPS index can validate the effectiveness of our domain-gap aware training clearly.
Some visual examples by different SRN settings are also included in Table 2.
By introducing target domain data in the training process of SRN, even the model trained with bicubic downsampled LR-HR pairs $\{\mathcal{Y}^b, \mathcal{X}^r\}$ generalize well on real-world LR images.

\vspace{2mm}
\noindent\textbf{Domain-distance weighted  Supervision.}
Besides the domain-gap aware training, we also proposed a domain-distance weighted supervision strategy to make better  use of source domain data.
The experimental results in Table 2 also clearly demonstrate the advantage of domain-distance weighted supervision.
By introducing weights to adaptively exploit paired training data, we are able to achieve better SRN over the baseline models.
In addition, the proposed two strategies are complementary, when combining the two strategies together, our final DASR method achieves significant improvement over the models which only adopts one of the two strategies.

\subsection{Comparison with State-of-the-Arts.}

\begin{figure}[t!]
      \vspace{-3mm}
        \scriptsize
        \centering
        \begin{tabular}{p{0.25\columnwidth}<{\centering}p{0.77\columnwidth}<{\centering}}
          \begin{adjustbox}{valign=t}
            \tiny
              \begin{tabular}{c}
                  \includegraphics[height=0.18\columnwidth,width=0.25\columnwidth]{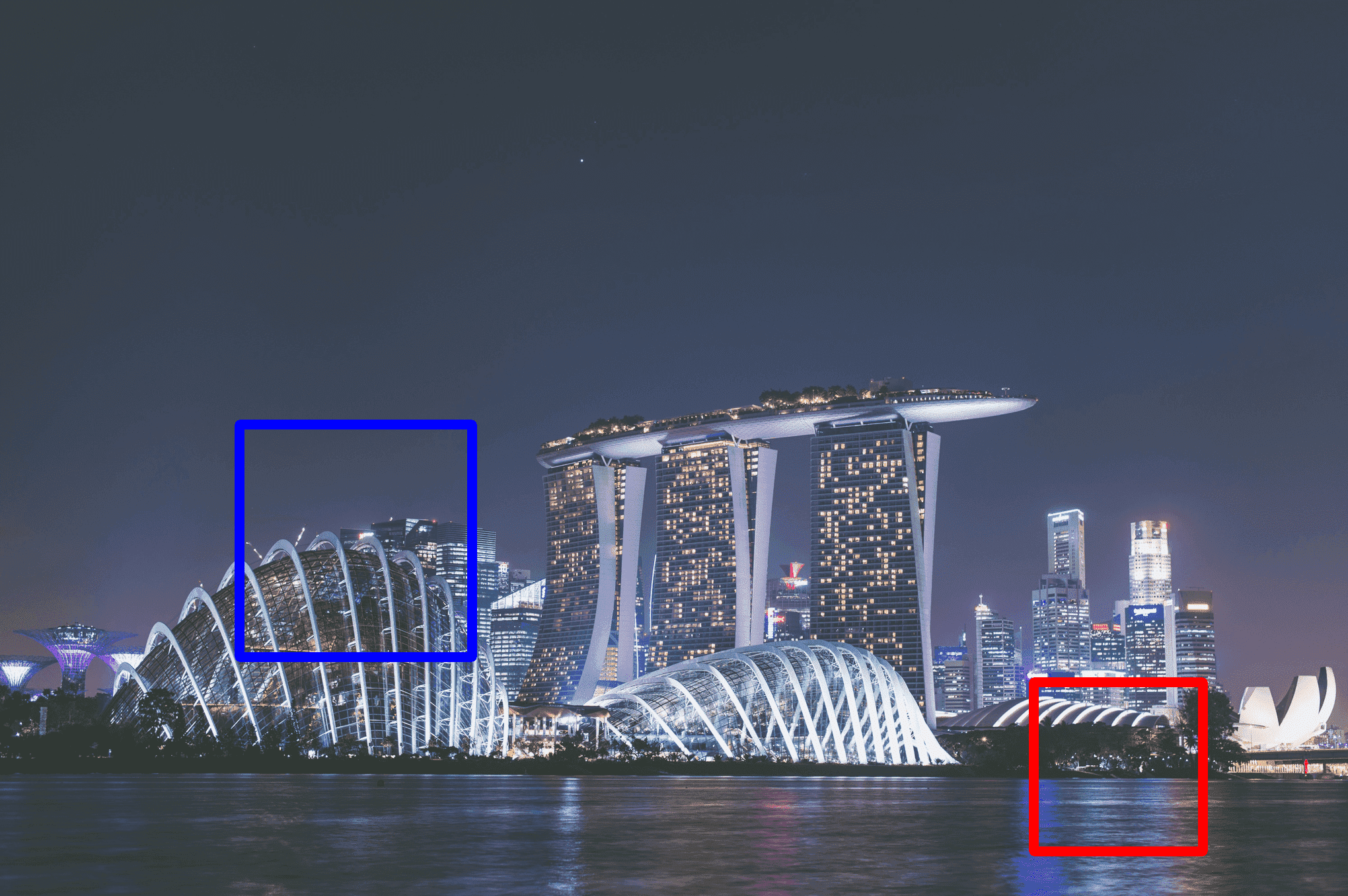}
                  \\
  
                  \begin{tabular}{cc}
                    \includegraphics[width=0.12\columnwidth]{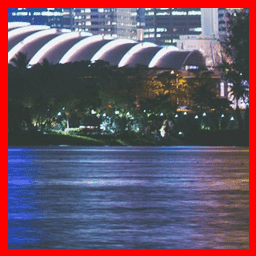} &
                    \includegraphics[width=0.12\columnwidth]{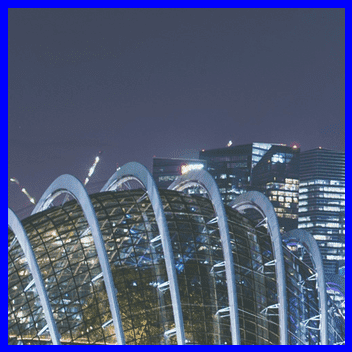}
                    
                  \end{tabular}
              \end{tabular}
          \end{adjustbox}
          &
        \hspace{-4mm}
        \begin{adjustbox}{valign=t}
        \tiny
        \centering
           \begin{tabular}{p{0.14\columnwidth}<{\centering}p{0.14\columnwidth}<{\centering}p{0.14\columnwidth}<{\centering}p{0.14\columnwidth}<{\centering}p{0.14\columnwidth}<{\centering}}
              \includegraphics[width=0.15\columnwidth]{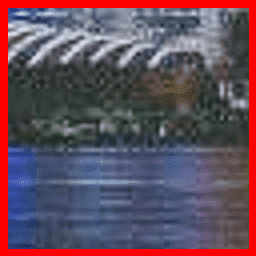} &
              \includegraphics[width=0.15\columnwidth]{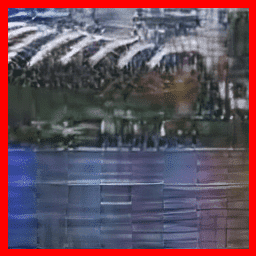} &
              \includegraphics[width=0.15\columnwidth]{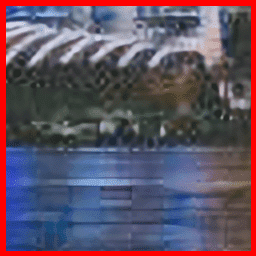} &
              \includegraphics[width=0.15\columnwidth]{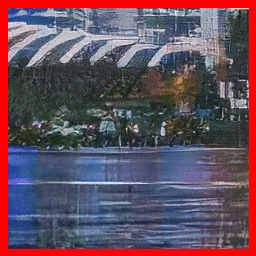}&
              \includegraphics[width=0.15\columnwidth]{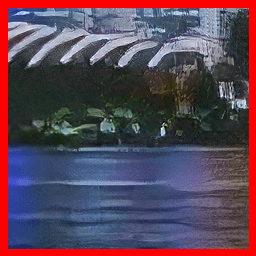}
              \\
              \includegraphics[width=0.15\columnwidth]{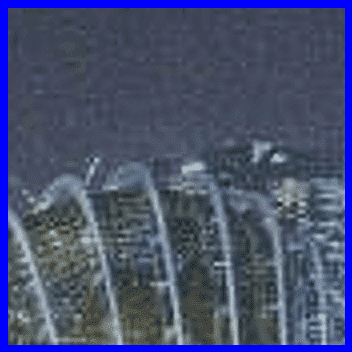} &
              \includegraphics[width=0.15\columnwidth]{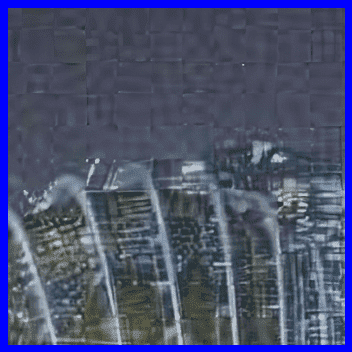} &
              \includegraphics[width=0.15\columnwidth]{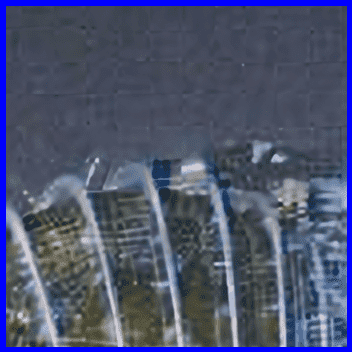} &
              \includegraphics[width=0.15\columnwidth]{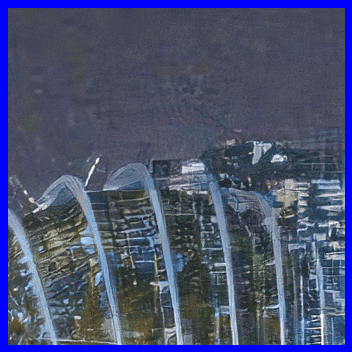} &
              \includegraphics[width=0.15\columnwidth]{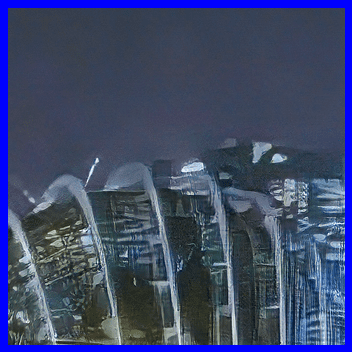}
           \end{tabular}
        \end{adjustbox}
      \\
      \\
      \begin{adjustbox}{valign=t}
        \tiny
          \begin{tabular}{c}
            \includegraphics[height=0.18\columnwidth,width=0.25\columnwidth]{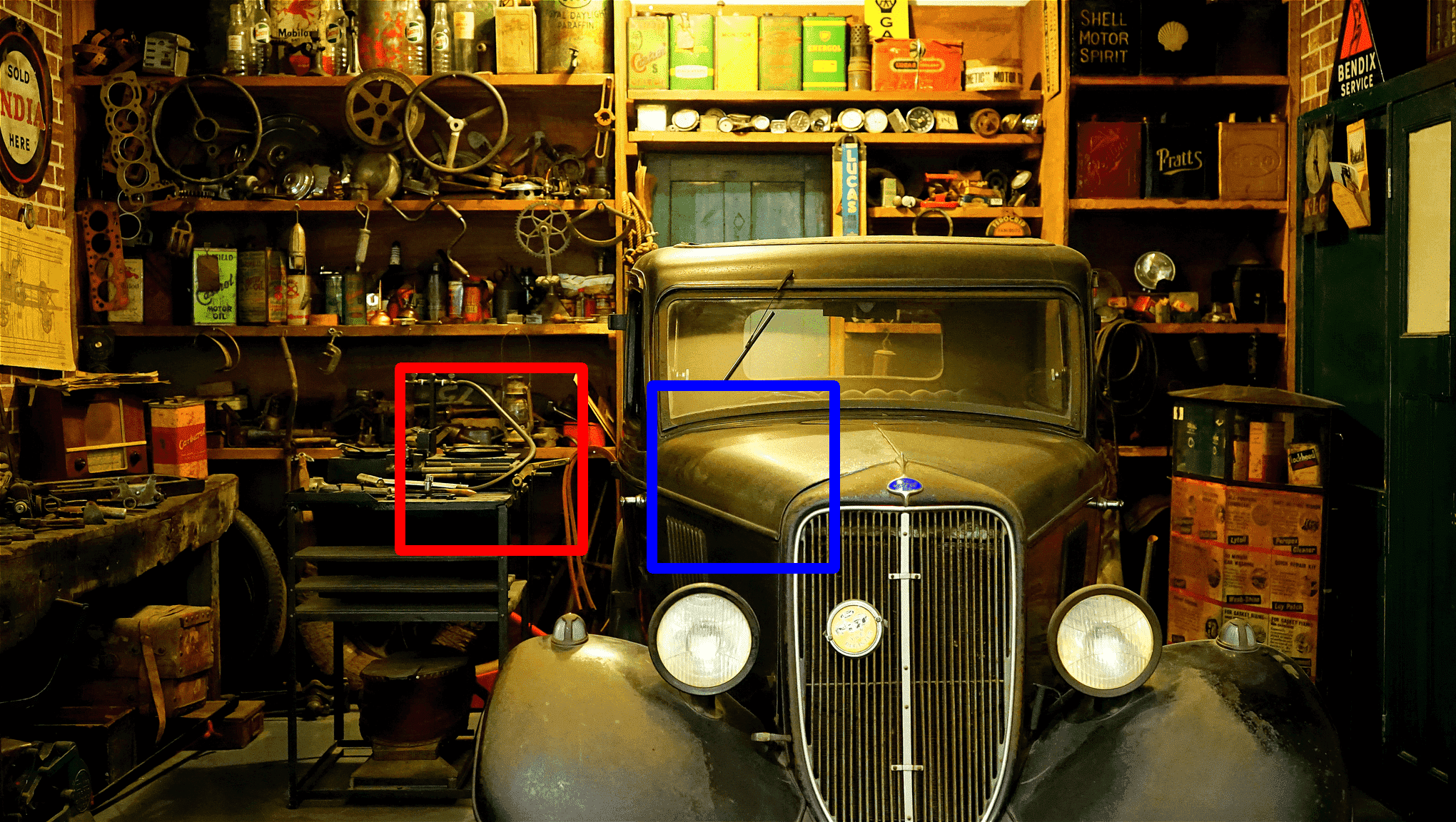}                
            \\
            \begin{tabular}{cc}
              \includegraphics[width=0.12\columnwidth]{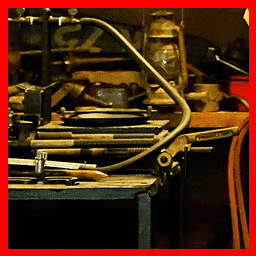} &
              \includegraphics[width=0.12\columnwidth]{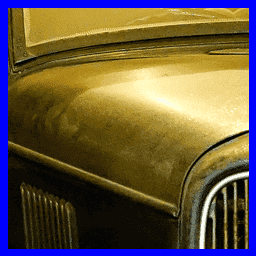}
            \end{tabular}
          \end{tabular}
      \end{adjustbox}
      &
      
      \hspace{-4mm}
      \begin{adjustbox}{valign=t}
      \tiny
      \centering
      \begin{tabular}{p{0.14\columnwidth}<{\centering}p{0.14\columnwidth}<{\centering}p{0.14\columnwidth}<{\centering}p{0.14\columnwidth}<{\centering}p{0.14\columnwidth}<{\centering}}
          \includegraphics[width=0.15\columnwidth]{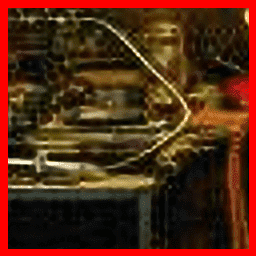} &
          \includegraphics[width=0.15\columnwidth]{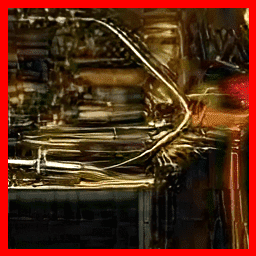} &
          \includegraphics[width=0.15\columnwidth]{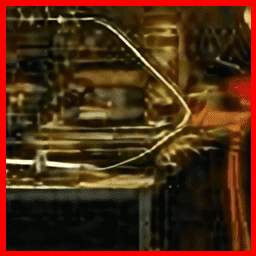} &
          \includegraphics[width=0.15\columnwidth]{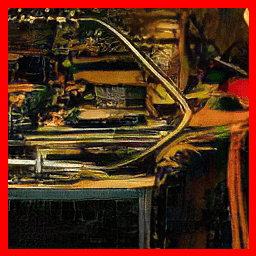}&
          \includegraphics[width=0.15\columnwidth]{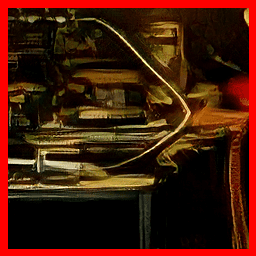}
          \\
          \includegraphics[width=0.15\columnwidth]{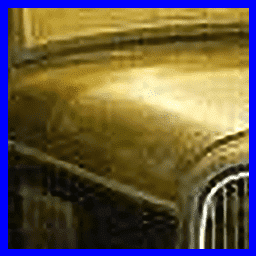} &
          \includegraphics[width=0.15\columnwidth]{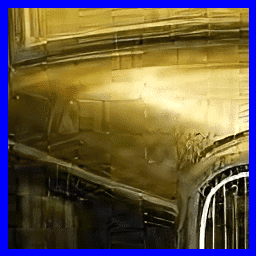} &
          \includegraphics[width=0.15\columnwidth]{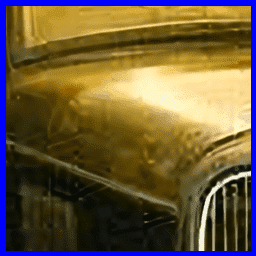} &
          \includegraphics[width=0.15\columnwidth]{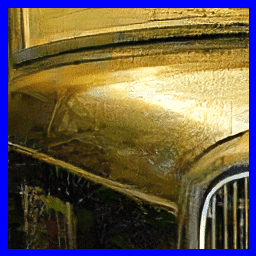} &
          \includegraphics[width=0.15\columnwidth]{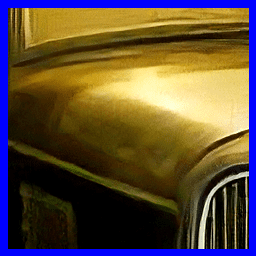}
       \end{tabular}
      \end{adjustbox}
      \\
      \\
    \begin{adjustbox}{valign=t}
      \tiny
        \begin{tabular}{c}
          \includegraphics[height=0.18\columnwidth,width=0.25\columnwidth]{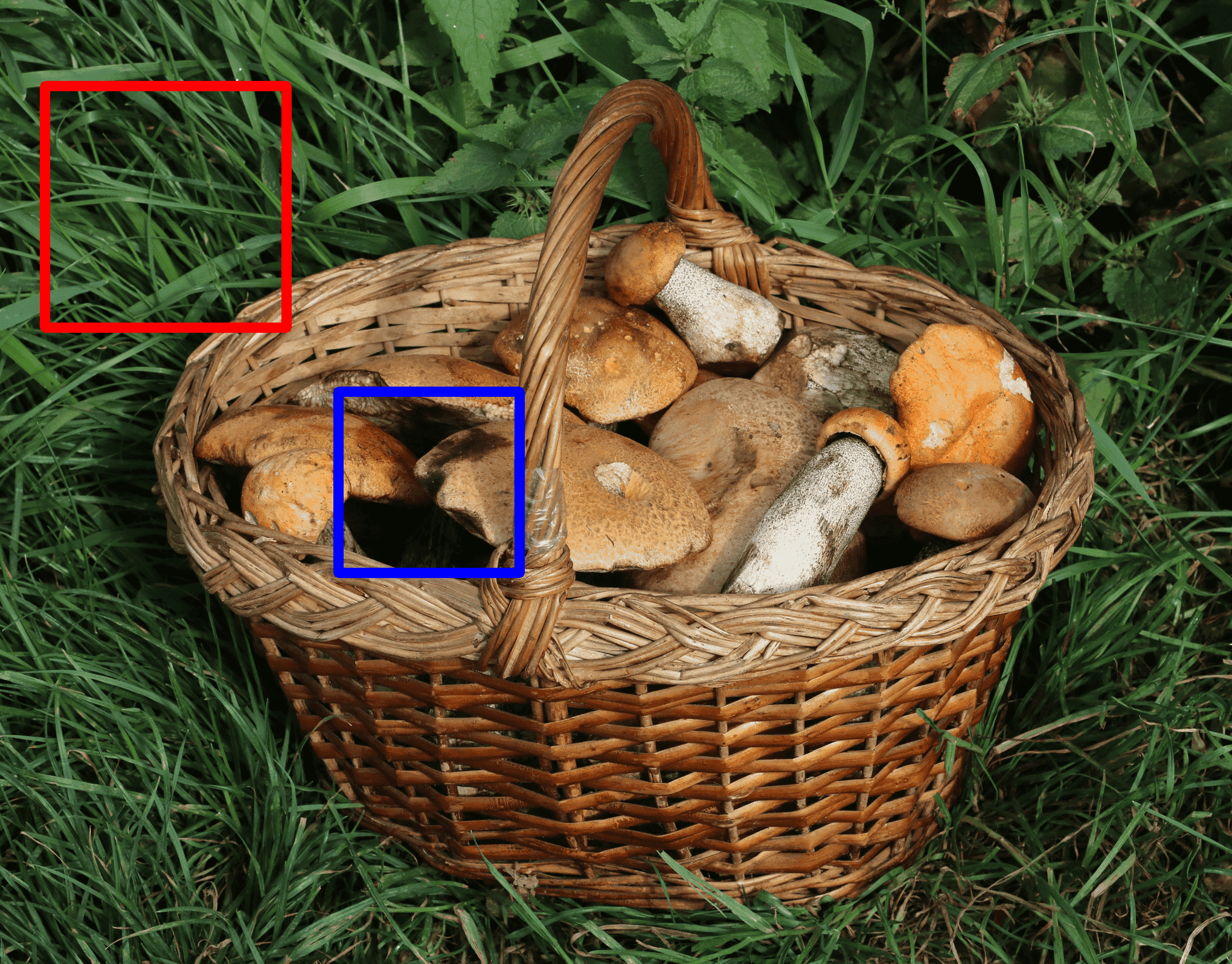}                
          \\
          \begin{tabular}{cc}
            \includegraphics[width=0.12\columnwidth]{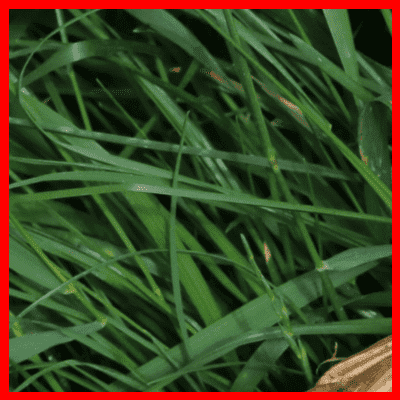} &
            \includegraphics[width=0.12\columnwidth]{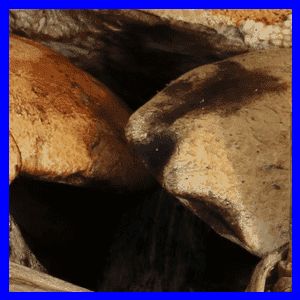}
          \end{tabular}
            \\
            Ground Truth
            \\
        \end{tabular}
    \end{adjustbox}
    &
    \hspace{-4mm}
    \begin{adjustbox}{valign=t}
    \tiny
    \centering
    \begin{tabular}{p{0.14\columnwidth}<{\centering}p{0.14\columnwidth}<{\centering}p{0.14\columnwidth}<{\centering}p{0.14\columnwidth}<{\centering}p{0.14\columnwidth}<{\centering}}
        \includegraphics[width=0.15\columnwidth]{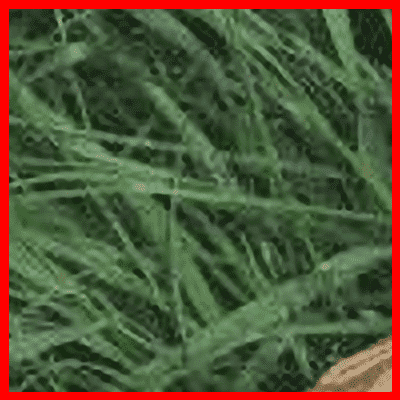} &
        \includegraphics[width=0.15\columnwidth]{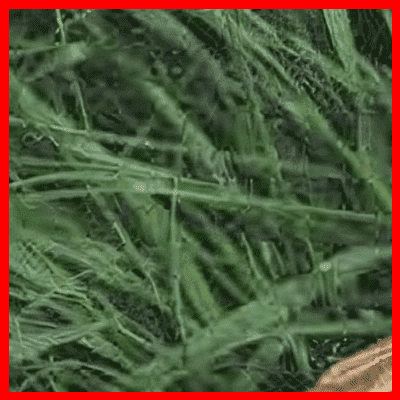}&
        \includegraphics[width=0.15\columnwidth]{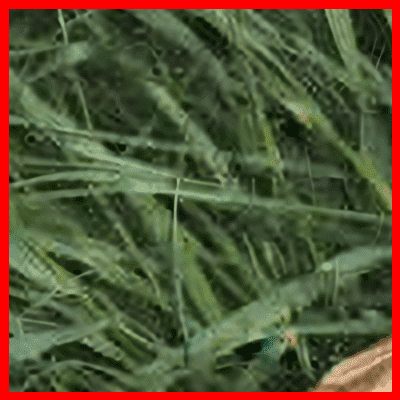} &
        \includegraphics[width=0.15\columnwidth]{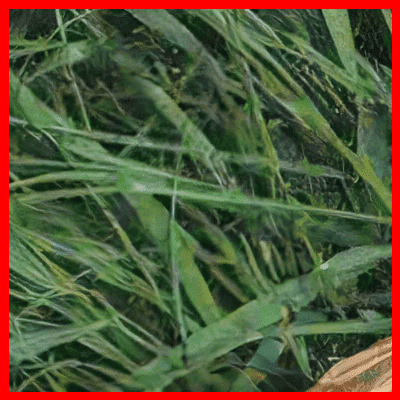}&
        \includegraphics[width=0.15\columnwidth]{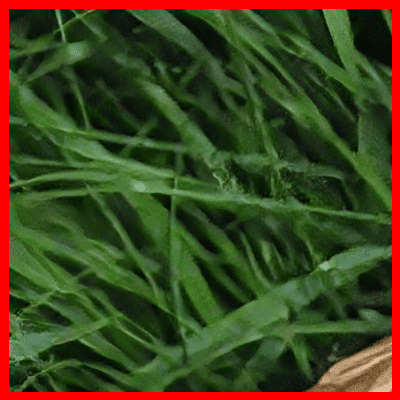}
        \\
        \includegraphics[width=0.15\columnwidth]{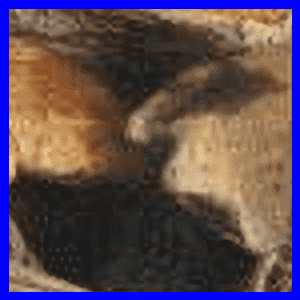} &
        \includegraphics[width=0.15\columnwidth]{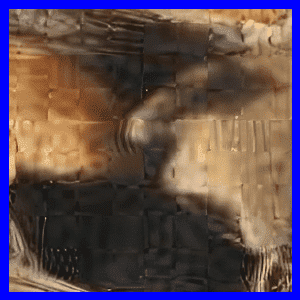} &
        \includegraphics[width=0.15\columnwidth]{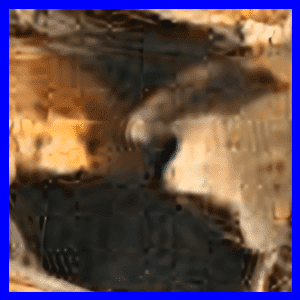} &
        \includegraphics[width=0.15\columnwidth]{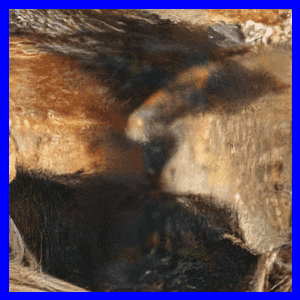} &
        \includegraphics[width=0.15\columnwidth]{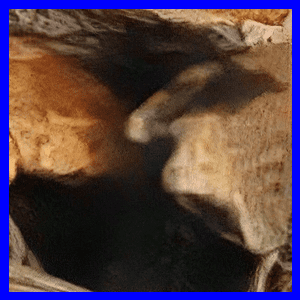}
        \\
        ZSSR~\cite{ZSSR}&
        P.T.&
        CinCGAN~\cite{Yuan2018UnsupervisedIS} &
        FSSR~\cite{Fritsche2019FrequencySF}&
        DASR 
        \\
        &ESRGAN~\cite{ESRGAN}&&&(ours)
       \end{tabular}
      \end{adjustbox}
        \end{tabular}
        \caption{SR results by different methods on testing images from AIM Challenge on Real World SR at ICCV 2019~\cite{lugmayr2019aim}.}
        
        \vspace{-3mm}
        \label{AIM_visual}
      \end{figure}

In this section, we compare our method with other real-world
super-resolution method. 
The competing approaches include Zero-shot SR (ZSSR) \cite{ZSSR} and unpaired approaches Frequency Separation for Super Resolution (FSSR) \cite{Fritsche2019FrequencySF} and cycle-in-cycle generative adversarial networks 
(CinCGAN) \cite{Yuan2018UnsupervisedIS}.
ZSSR applies a Zero-Shot learning strategy in the testing phase to adapt to image-specific degradation model.
CinCGAN and FSSR are recently proposed unsupervised  SR approach, FSSR is the winner of the AIM Challenge on Real World SR at ICCV 2019~\cite{lugmayr2019aim}.
The code of FSSR \cite{Fritsche2019FrequencySF} is provided by the paper authors, and CinCGAN model \cite{Yuan2018UnsupervisedIS} is implemented by ourselves.
In addition, we also provide the results by pre-trained ESRGAN (denote as P.T. ESRGAN) for reference, the pre-trained ESRGAN model was trained on synthetic dataset with bicubic downsampled LR images.
The quantitative metrics achieved by different methods are shown in Table \ref{table:resutls}.
In Fig. \ref{AIM_visual}, we also provide some visual examples of the SR results by different approaches.
The quantitative metrics as well as visual examples clearly demonstrate that our proposed DASR approach is superior to the competing models.
The degradation assumptions by ZSSR and Pre-trained ESRGAN  can not reflect the complex degradation adopted in the AIM challenge, both the two approaches generate strange artifacts in the HR estimation.
While, the FSSR approach generates better synthetic data which have similar characteristic with real-world image to train the model, is able to deliver better SR results than the ZSSR and pre-trained ESRGAN approach.
But FSSR does not consider the domain gap between generated and real LR images, still creates artifacts in the final output.
Our novel DASR exploits information in the target domain in the training phase, is able to generate high quality SR estimations which have visually pleasant textures and less artifacts.
More visual examples can be found in our supplementary material.

\begin{table}[!t]
    \vspace{-3mm}
    \caption{Quantitative comparison on different datasets. More experimental details can be found in section 4 and 5. Please note that supervisely trained ESRGAN (S.T. ESRGAN) is trained with paired training data while the other methods are trained without paired training data.}
    \label{table:resutls}
    \begin{center}

    \scriptsize
    \begin{tabular}{l|cccc|cccc|cccc}
    \hline
    &\multicolumn{4}{c|}{AIM~\cite{lugmayr2019aim}}&\multicolumn{4}{c|}{RealSR~\cite{cai2019toward}}&\multicolumn{4}{c}{CameraSR~\cite{CameraSR}}\\
  
    Methods&  PSNR & SSIM& LPIPS& MOS&
              PSNR & SSIM& LPIPS& MOS&
              PSNR & SSIM& LPIPS& MOS\\
  
    \hline
  
    ZSSR  &22.327 &0.6022&0.630&3.10&26.007&0.7482&0.386&3.42&-&-&-&-\\
    P.T. ESRGAN& 21.382 &0.5478&0.543&3.16&25.956&0.7468&0.415&3.08&-&-&-&-\\
    CinCGAN &21.602&0.6129&0.461&3.56&25.094&0.7459&0.405&3.24&-&-&-&- \\
    FSSR & 20.820 &0.5103&0.390&2.41&25.992&0.7388&0.265&2.45&23.781&0.7566&0.180&3.14\\
    DASR(ours) & 21.600 &0.5640&0.336&1.94&26.782&0.7822&0.228&2.05&25.769&0.8312&0.151&2.60\\
    \hline
    S.T. ESRGAN&-&-&-&-&25.704&0.7487&0.199&1.35&25.346&0.8036&0.111&1.18\\
    \hline
    \end{tabular}
    \end{center}
    \vspace{-5mm}
\end{table}
    
   \vspace{-3mm}
\section{Experimental Results on Real-World Images}
In this section, we evaluate the proposed DASR model on real-world datasets.
We conduct experiments on two real-world image SR datasets: RealSR \cite{cai2019toward} and CameraSR \cite{CameraSR}.
The two datasets contain real LR-HR pairs collected by adjusting the focal length of digital cameras.
We adopt the LR images in the two datasets and HR images in the DIV2K \cite{Timofte_2018_CVPR_Workshops} dataset to deploy our unsupervised training, and evaluate our models on the validation sets which have paired data for quantitative evaluation.

     \vspace{-2mm}

\subsection{Experimental Results on RealSR Dataset}
\begin{figure}[!hbt]
\scriptsize
\centering
\begin{tabular}{cc}
  \begin{adjustbox}{valign=t}
    \tiny
      \begin{tabular}{c}
          \includegraphics[height=0.16\columnwidth,width=0.21\columnwidth]{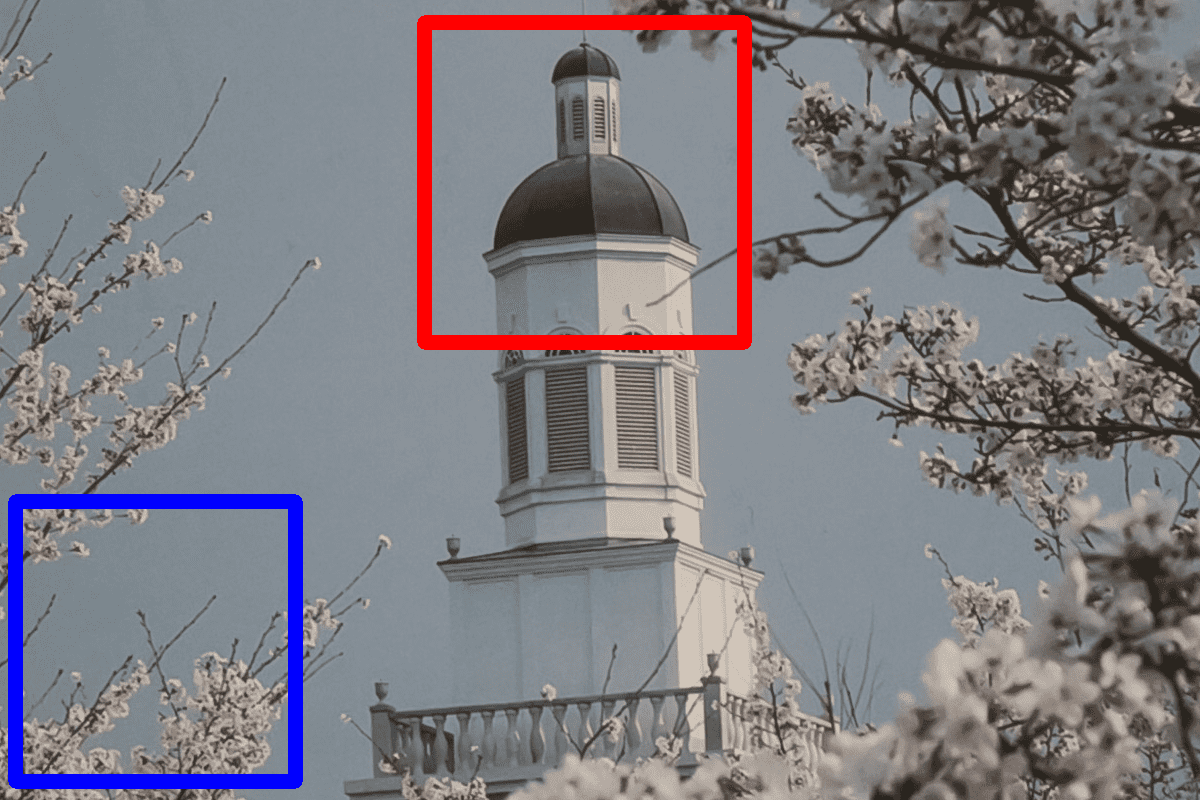}

          \\

          \begin{tabular}{cc}
            \includegraphics[width=0.1\columnwidth]{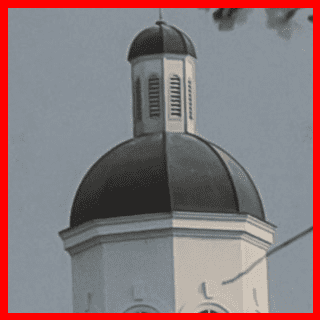} &
            \includegraphics[width=0.1\columnwidth]{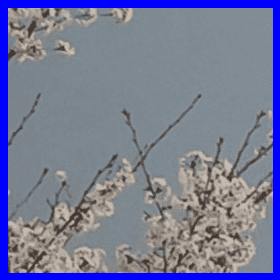} 
            
          \end{tabular}
      \end{tabular}
  \end{adjustbox}
  &
\hspace{-2mm}
\begin{adjustbox}{valign=t}
\tiny
\centering
   \begin{tabular}{p{0.12\columnwidth}<{\centering}p{0.12\columnwidth}<{\centering}p{0.12\columnwidth}<{\centering}p{0.12\columnwidth}<{\centering}p{0.12\columnwidth}<{\centering}p{0.12\columnwidth}<{\centering}}
      \includegraphics[width=0.13\columnwidth]{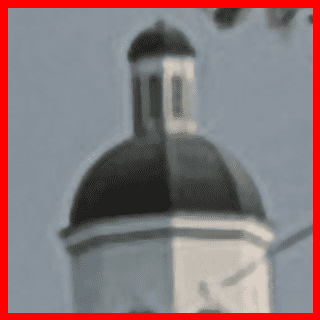} &
      \includegraphics[width=0.13\columnwidth]{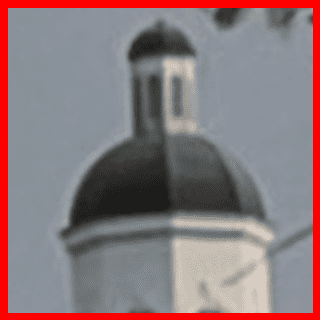}&
      \includegraphics[width=0.13\columnwidth]{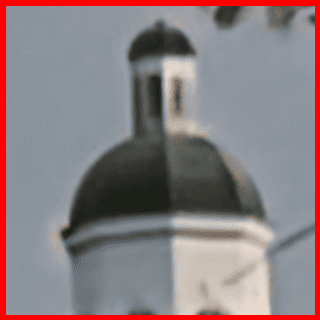}&
      \includegraphics[width=0.13\columnwidth]{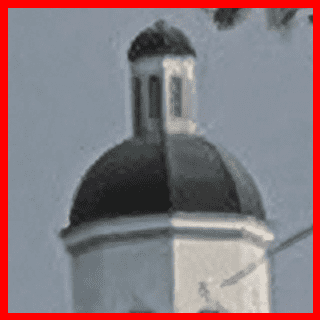}&
      \includegraphics[width=0.13\columnwidth]{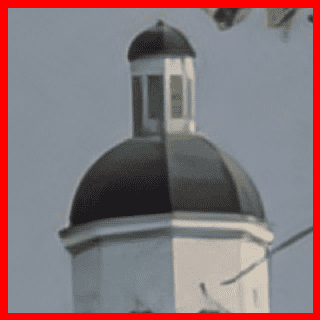}&
      \includegraphics[width=0.13\columnwidth]{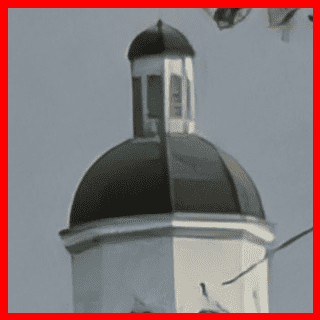}
      \\
      \includegraphics[width=0.13\columnwidth]{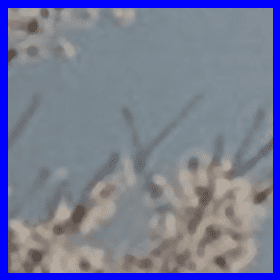} &
      \includegraphics[width=0.13\columnwidth]{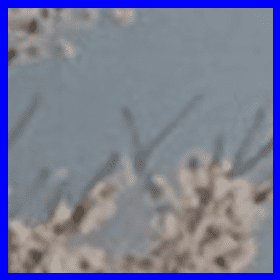}  &
      \includegraphics[width=0.13\columnwidth]{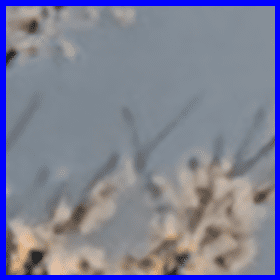}  &
      \includegraphics[width=0.13\columnwidth]{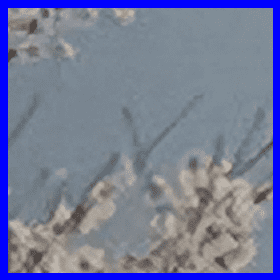}  &
      \includegraphics[width=0.13\columnwidth]{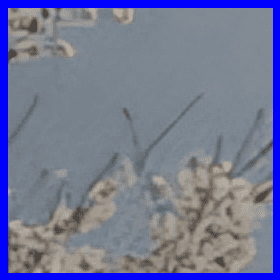} &
      \includegraphics[width=0.13\columnwidth]{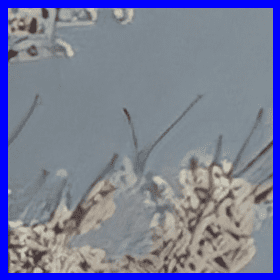} 
   \end{tabular}
\end{adjustbox}

 \\
 \\

 \begin{adjustbox}{valign=t}
  \tiny
    \begin{tabular}{c}
        \includegraphics[height=0.16\columnwidth,width=0.21\columnwidth]{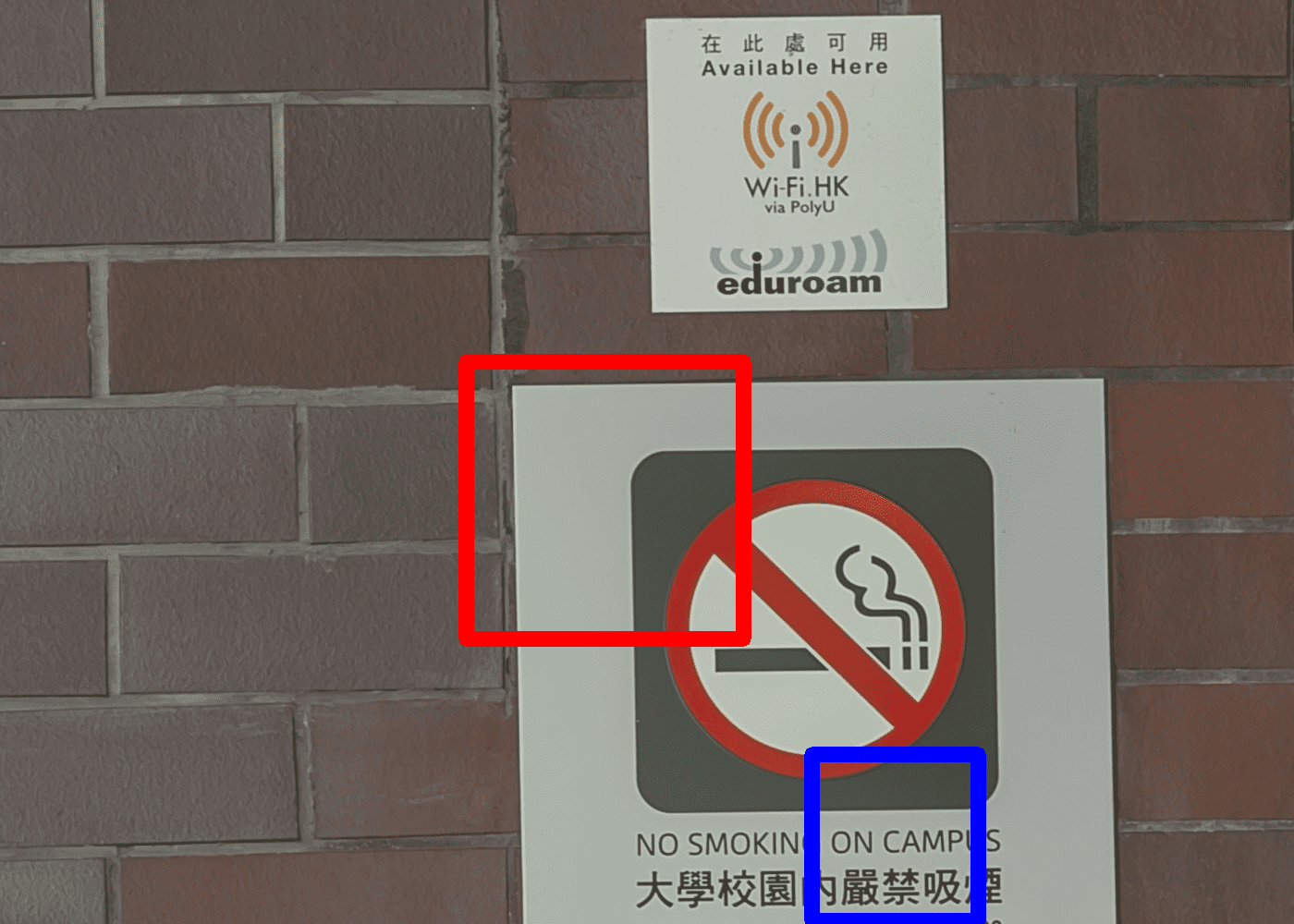}

        \\

        \begin{tabular}{cc}
          \includegraphics[width=0.1\columnwidth]{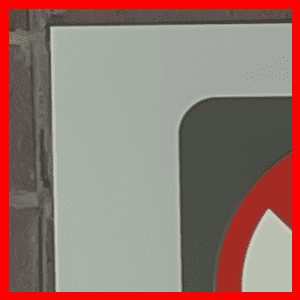} &
          \includegraphics[width=0.1\columnwidth]{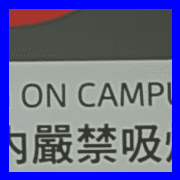}
          
        \end{tabular}
        
    \end{tabular}
\end{adjustbox}
&
\hspace{-2mm}
\begin{adjustbox}{valign=t}
\tiny
\centering
 \begin{tabular}{p{0.12\columnwidth}<{\centering}p{0.12\columnwidth}<{\centering}p{0.12\columnwidth}<{\centering}p{0.12\columnwidth}<{\centering}p{0.12\columnwidth}<{\centering}p{0.12\columnwidth}<{\centering}}
  \includegraphics[width=0.13\columnwidth]{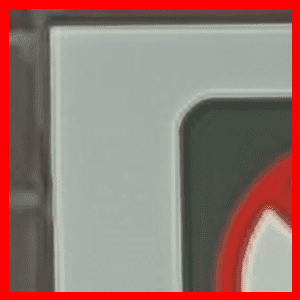} &
  \includegraphics[width=0.13\columnwidth]{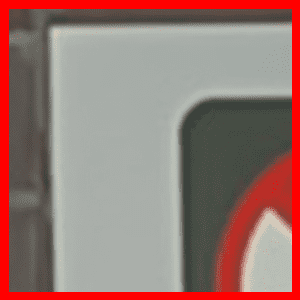}&
  \includegraphics[width=0.13\columnwidth]{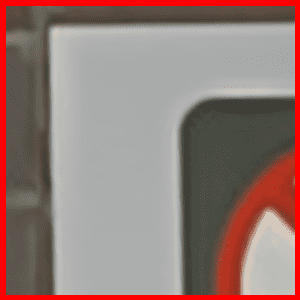}&
  \includegraphics[width=0.13\columnwidth]{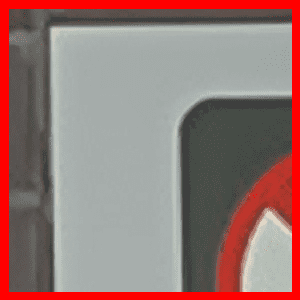}&
  \includegraphics[width=0.13\columnwidth]{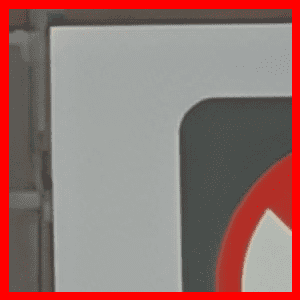}&
  \includegraphics[width=0.13\columnwidth]{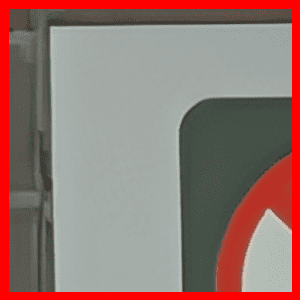}
  \\
  \includegraphics[width=0.13\columnwidth]{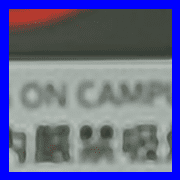} &
  \includegraphics[width=0.13\columnwidth]{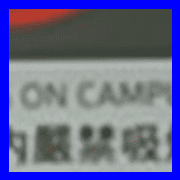}  &
  \includegraphics[width=0.13\columnwidth]{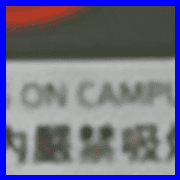}  &
  \includegraphics[width=0.13\columnwidth]{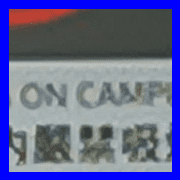} &
  \includegraphics[width=0.13\columnwidth]{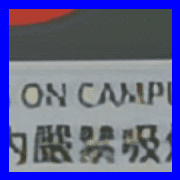}&
  \includegraphics[width=0.13\columnwidth]{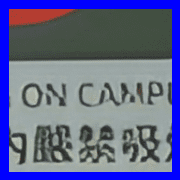} 
 \end{tabular}
\end{adjustbox}
\\
\\
      \begin{adjustbox}{valign=t}
        \tiny
          \begin{tabular}{c}
              \includegraphics[height=0.16\columnwidth,width=0.21\columnwidth]{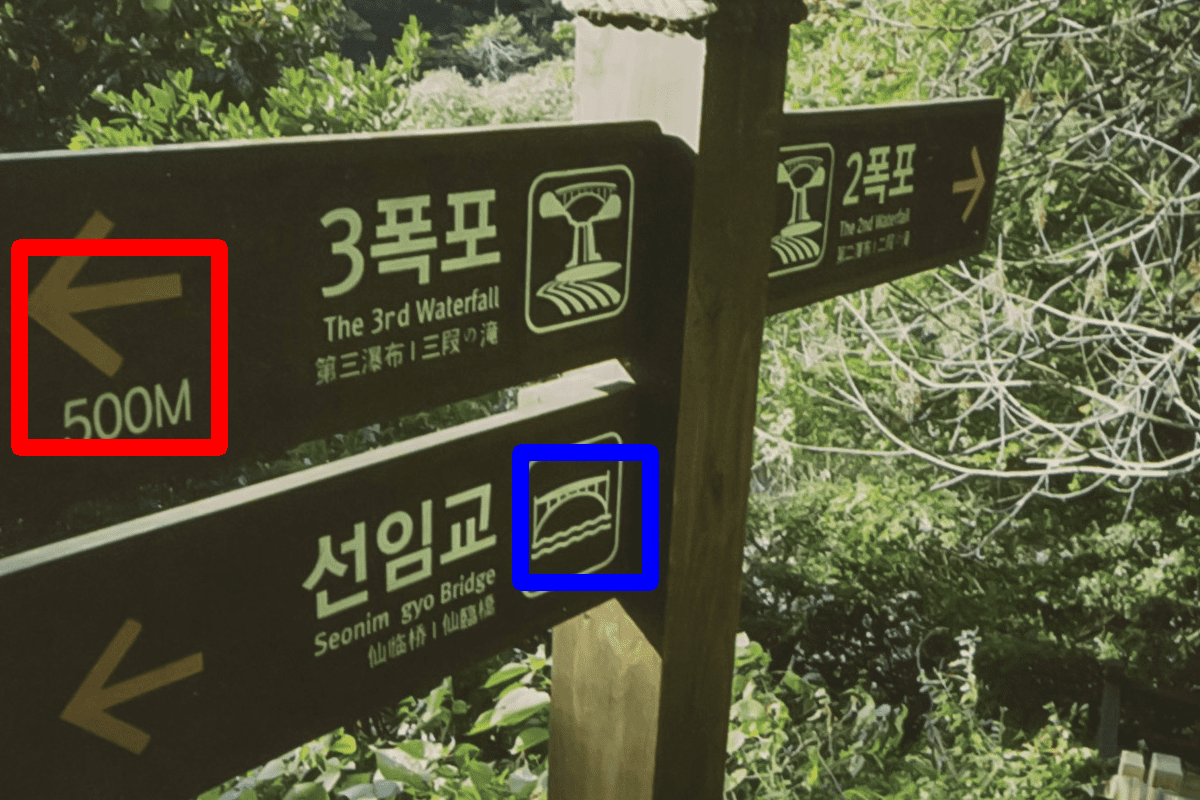}
      
              \\
      
              \begin{tabular}{cc}
                \includegraphics[width=0.1\columnwidth]{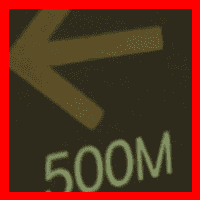}&
                \includegraphics[width=0.1\columnwidth]{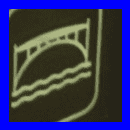}
                
              \end{tabular}

                \\
                Ground Truth
              
          \end{tabular}
      \end{adjustbox}
      &
      \hspace{-2mm}
      \begin{adjustbox}{valign=t}
      \tiny
      \centering
       \begin{tabular}{p{0.12\columnwidth}<{\centering}p{0.12\columnwidth}<{\centering}p{0.12\columnwidth}<{\centering}p{0.12\columnwidth}<{\centering}p{0.12\columnwidth}<{\centering}p{0.12\columnwidth}<{\centering}}
        \includegraphics[width=0.13\columnwidth]{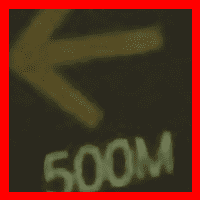} &
        \includegraphics[width=0.13\columnwidth]{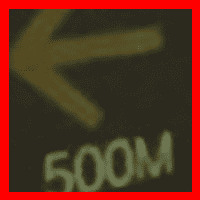}  &
        \includegraphics[width=0.13\columnwidth]{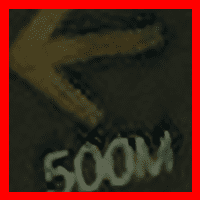}  &
        \includegraphics[width=0.13\columnwidth]{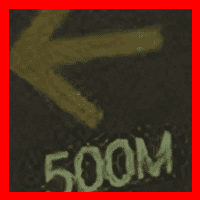} &
        \includegraphics[width=0.13\columnwidth]{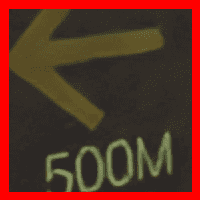}&
        \includegraphics[width=0.13\columnwidth]{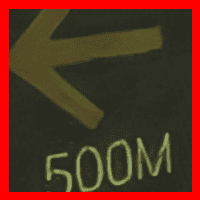} 
        \\
        \includegraphics[width=0.13\columnwidth]{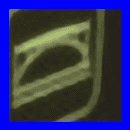} &
        \includegraphics[width=0.13\columnwidth]{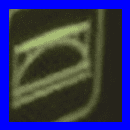}&
        \includegraphics[width=0.13\columnwidth]{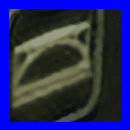}&
        \includegraphics[width=0.13\columnwidth]{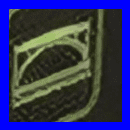}&
        \includegraphics[width=0.13\columnwidth]{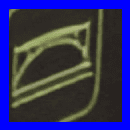}&
        \includegraphics[width=0.13\columnwidth]{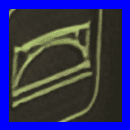}
        \\
        ZSSR\cite{ZSSR}&
        P.T.&
        CinCGAN&
        FSSR\cite{Fritsche2019FrequencySF}&
        DASR &
        S.T.
        \\
        &ESRGAN\cite{ESRGAN}&~\cite{Yuan2018UnsupervisedIS}&&(ours)&ESRGAN\cite{ESRGAN}
 \end{tabular}
\end{adjustbox}

\end{tabular}
  \vspace{-2mm}
\caption{SR results by different methods on testing images from RealSR 
\cite{cai2019toward}.}
    \vspace{-2mm}
\label{Real_visual}
\end{figure}

\vspace{-2mm}
RealSR \cite{cai2019toward} is a recently collected real-world SR dataset.
The authors  utilize a Canon and a Nikon camera to collect 595 real LR-HR pairs by adjusting the focal length of the cameras, and adopt image registration algorithm to achieve aligned image pairs.
In our experiments, we utilize the 200 LR images collected by the Canon camera as our real-world LR images, and the 800 HR images in the DIV2K \cite{Timofte_2018_CVPR_Workshops} as our  HR images.
We train our DASR model as well as FSSR \cite{Fritsche2019FrequencySF} and CinCGAN \cite{Yuan2018UnsupervisedIS} models with the same data.
After unsupervised training, we employ our model to super-resolve LR images in the validation set of RealSR \cite{cai2019toward}, which consists of 100 LR-HR pairs.
The SR results generated by our model and the competing approaches are shown in Table \ref{table:resutls}.
Besides the ZSSR \cite{ZSSR}, FSSR \cite{Fritsche2019FrequencySF} and pre-trained ESRGAN, we also provide the results by supervisely trained ESRGAN (denote as S.T. ESRGAN) for reference, which utilizes the real paired data in the training set to train the ESRGAN model in a fully supervised manner.
DASR significantly outperforms other
blind super-resolution methods in both LPIPS index and MOS score.
Compared with the Supervised ESRGAN, DASR achieves comparable LPIPS indexes.
Some visual examples by different approaches are shown in Fig. \ref{Real_visual}, more visual examples can be found in our supplementary file.

\subsection{Experimental Results on CameraSR}
\begin{figure}[!hbt]

  \scriptsize
  \centering
  \begin{tabular}{cc}
    \begin{adjustbox}{valign=t}
      \tiny
        \begin{tabular}{c}
            \includegraphics[height=0.285\columnwidth,width=0.45\columnwidth]{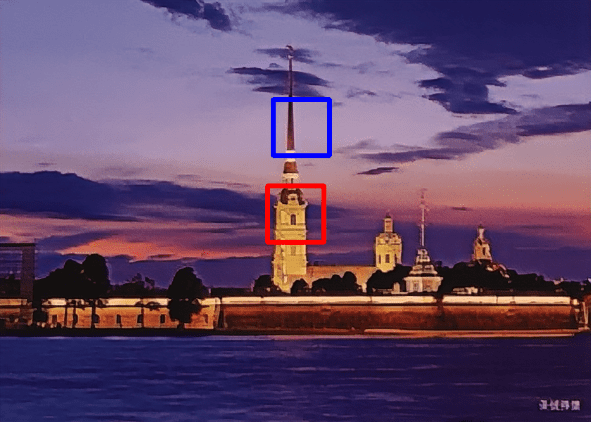}          
        \end{tabular}
    \end{adjustbox}
    &
  \hspace{-1mm}
  \begin{adjustbox}{valign=t}
  \tiny
  \centering
     \begin{tabular}{p{0.12\columnwidth}<{\centering}p{0.12\columnwidth}<{\centering}p{0.12\columnwidth}<{\centering}p{0.12\columnwidth}<{\centering}}
        \includegraphics[width=0.14\columnwidth]{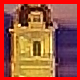} &
        \includegraphics[width=0.14\columnwidth]{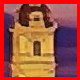}&
        \includegraphics[width=0.14\columnwidth]{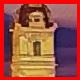}&
        \includegraphics[width=0.14\columnwidth]{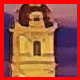}
        \\
        \includegraphics[width=0.14\columnwidth]{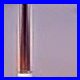} &
        \includegraphics[width=0.14\columnwidth]{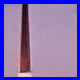} &
        \includegraphics[width=0.14\columnwidth]{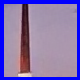}&
        \includegraphics[width=0.14\columnwidth]{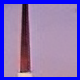}
     \end{tabular}
  \end{adjustbox}

  \\
  \\

  \begin{adjustbox}{valign=t}
    \tiny
      \begin{tabular}{c}
          \includegraphics[height=0.285\columnwidth,width=0.45\columnwidth]{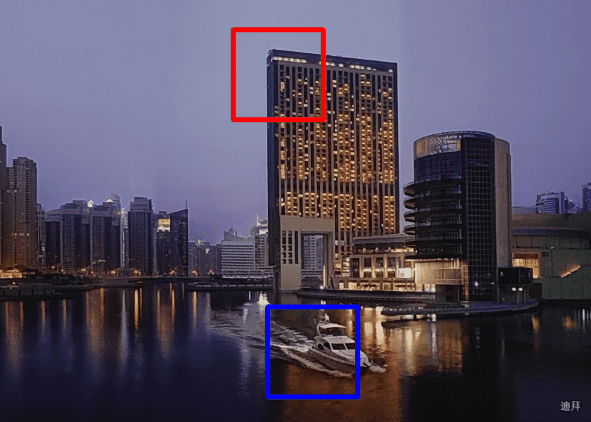}
          \\
          Ground Truth
          
      \end{tabular}
  \end{adjustbox}
  &
\hspace{-1mm}
\begin{adjustbox}{valign=t}
\tiny
\centering
   \begin{tabular}{p{0.12\columnwidth}<{\centering}p{0.12\columnwidth}<{\centering}p{0.12\columnwidth}<{\centering}p{0.12\columnwidth}<{\centering}}
      \includegraphics[width=0.14\columnwidth]{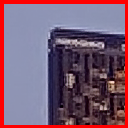} &
      \includegraphics[width=0.14\columnwidth]{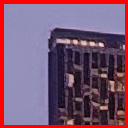}&
      \includegraphics[width=0.14\columnwidth]{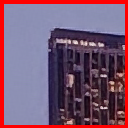}&
      \includegraphics[width=0.14\columnwidth]{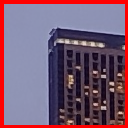}
      \\
      \includegraphics[width=0.14\columnwidth]{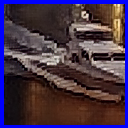} &
      \includegraphics[width=0.14\columnwidth]{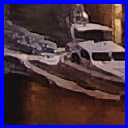} &
      \includegraphics[width=0.14\columnwidth]{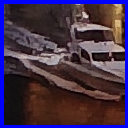}&
      \includegraphics[width=0.14\columnwidth]{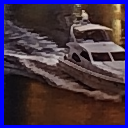}
      \\
      FSSR\cite{Fritsche2019FrequencySF}&
      DASR&
      S.T.&
      HR
      \\
      &(ours)&ESRGAN\cite{ESRGAN}&
   \end{tabular}
\end{adjustbox}
\end{tabular}

        \vspace{-2mm}
  \caption{SR results by different methods on a testing image from CameraSR \cite{CameraSR}.}
  \label{Camera_visual}
        \vspace{-2mm}
\end{figure}

We also compare different approaches on the CameraSR \cite{CameraSR} dataset.
CameraSR contains 100 LR-HR pairs captured by an iphoneX and a Nikon Camera, respectively.
We test our method on the iphoneX subset.
As the LR and HR images in the dataset are with the same spatial size, we remove the down-sampling and up-sampling operations in our framework as well as the FSSR model.
Similar to our experiments on the RealSR dataset, we utilize the 100 LR images in CameraSR training set and 800 HR images in the DIV2K \cite{Timofte_2018_CVPR_Workshops} to train our model as well as FSSR.
The SR results by our model and FSSR are shown in Table \ref{table:resutls}.
DASR outperform FSSR by a large margin. 
Visual examples are shown in Fig. \ref{Camera_visual}, more results can be found in our supplementary file.

      \vspace{-2mm}

\section{Conclusions}

We propose a novel DASR framework for unsupervised real-world image SR.
Given only unpaired data, DASR firstly trains a down-sampling network to generate synthetic LR images  
 in the real-world LR distribution.
Then, the generated synthetic pairs and real LR images are exploited to train the SR network under a domain adaptation setting.
We proposed a domain-gap aware training strategy to introduce adversarial loss in the target domain, and a domain-distance weighted supervision strategy to take better advantage of synthetic data in the source domain.
Our experimental results on synthetic and real-world datasets demonstrate the  effectiveness of our approach for real-world SR.

\clearpage
%
%
\bibliographystyle{splncs04}
\bibliography{ms}
\end{document}